\documentclass{article}

\newcommand{\eat}[1]{}

\newtheorem{example}{Example}

\usepackage[ruled,linesnumbered]{algorithm2e}
\usepackage{algpseudocode}
\usepackage{forest}
\usepackage{adjustbox} 
\usepackage{makecell}      
\usepackage{colortbl}      
\usepackage{xcolor}        
\usepackage{pifont}        
\usepackage{booktabs}
\usepackage{multirow}
\usepackage{graphicx}  
\usepackage{pifont} 
\usepackage{fontspec}
\usepackage{xeCJK}  

\usepackage{fontawesome5} 
\usepackage{wrapfig}
\usepackage{tabularx}
\usepackage{multirow}
\usepackage{dblfloatfix}

\algnewcommand\algorithmicinput{\textbf{Input:}}
\algnewcommand\algorithmicoutput{\textbf{Output:}}
\algnewcommand\Input{\item[\algorithmicinput]}
\algnewcommand\Output{\item[\algorithmicoutput]}

\algdef{SE}[DOWHILE]{Do}{doWhile}{\algorithmicdo}[1]{\algorithmicwhile\ #1}%
\usepackage{lineno}

\definecolor{green}{RGB}{0,128,0}

\definecolor{yellow}{RGB}{255,200,18}

\newcommand{\stab}{\vspace{1.2ex}\noindent}

\newcommand{\labelagent}{\faTags}
\newcommand{\refineragent}{\faMagic}
\newcommand{\serializeragent}{\faStream}
\newcommand{\generatoragent}{\faEdit}
\newcommand{\validatoragent}{\faTools}

\newcommand{\bi}{\begin{itemize}}
\newcommand{\ei}{\end{itemize}}

\newcommand{\be}{\begin{enumerate}}
\newcommand{\ee}{\end{enumerate}}
\newcommand{\beqn}{\begin{eqnarray*}}
\newcommand{\eeqn}{\end{eqnarray*}}

\newcommand{\stitle}[1]{\stab\noindent{\bf #1}}

\newcommand{\eg}{{\em e.g.,}\xspace}

\newcommand{\eop}{\hspace*{\fill}\mbox{$\Box$}}     

\usepackage{tabu}                      
\usepackage{booktabs}                  
\usepackage{lipsum}                    
\usepackage{mwe}                       

\usepackage{newtxtext,newtxmath}

\usepackage{listings}
\usepackage{pifont}
\usepackage{tikz}
\usepackage{enumitem}
\usepackage{natbib}
\usepackage{utfsym} 
\usepackage[most]{tcolorbox}
\tcbuselibrary{skins, raster} 
\usetikzlibrary{shapes,snakes}
\usetikzlibrary{calc}

\definecolor{shadecolor}{RGB}{220,220,220}

\tikzstyle{mybox} = [draw=black, fill=black!5, thick,
    rectangle, rounded corners, inner sep=2pt, inner ysep=8pt]
\tikzstyle{fancytitle} =[fill=black, text=white]

\tcbset{
  stagebox/.style 2 args={
    enhanced,
    on line,
    valign=center,
    colback=#1,
    coltext=#2,
    arc=2pt,
    boxrule=0pt,
    boxsep=0pt,
    left=1pt,
    right=1pt,
    top=1.25pt,
    bottom=1pt,
    fontupper=\sffamily\bfseries\footnotesize,
  }
}

\newcommand{\doctable}{Doc2Table\xspace}
\newcommand{\docdb}{Doc2DB\xspace}
\newcommand{\dbdoc}{DB2Doc\xspace}
\newcommand{\ours}{Doc2DB-Bench\xspace}

\definecolor{mygreen}{RGB}{203,230,204}
\definecolor{mydarkgreen}{RGB}{100,230,100}
\definecolor{myred}{RGB}{255,182,193}
\definecolor{mygray}{RGB}{230,230,230}
\definecolor{myblue}{RGB}{138,180,189}
\definecolor{myyellow}{RGB}{255,185,84}
\definecolor{radargreen}{HTML}{38ADA9}

\definecolor{stageAcolor}{HTML}{90b3bd}
\definecolor{stageBcolor}{HTML}{738348}

\usepackage{xcolor}

\definecolor{DeltaUpBg}{HTML}{E6F4EA}
\definecolor{DeltaUpFg}{HTML}{137333}
\definecolor{DeltaDnBg}{HTML}{FDE7E9}
\definecolor{DeltaDnFg}{HTML}{A50E0E}
\definecolor{greyblue}{RGB}{208,220,232}

\definecolor{firstBest}{rgb}{0.86, 1, 0.86} 

\definecolor{myDarkGreen}{RGB}{0, 100, 0}
\usepackage{pifont}
\newcommand{\cmark}
{\textcolor{myDarkGreen}{\ding{51}}}%
\newcommand{\xmark}{\textcolor{red}{\ding{55}}}%

\newtcolorbox{findingbox}[2][]{
  colback=gray!10!white,
  colframe=gray!30!white,
  boxrule=0.2mm,
  left=0mm,
  right=0mm,
  top=0mm,
  bottom=0mm,
  coltitle=red!70!black,
  title={#2},
  #1
}

\newcommand{\ApplyGradient}[1]{%
    \pgfmathsetmacro{\percent}{#1}%
    \ifdim \percent pt > 100pt \def\percent{100}\fi
    \ifdim \percent pt < 0pt \def\percent{0}\fi
    \edef\col{\noexpand\cellcolor{blue!\percent!red!50}}%
    \col #1%
}

\newcolumntype{H}{>{\collectcell\ApplyGradient}c<{\endcollectcell}}

 \usepackage[preprint]{arxiv_2026}

\usepackage[utf8]{inputenc} 
\usepackage[T1]{fontenc}    
\usepackage{hyperref}       
\usepackage{url}            
\usepackage{booktabs}       
\usepackage{amsfonts}       
\usepackage{nicefrac}       
\usepackage{microtype}      
\usepackage{xcolor}         

\title{Doc2DB-Bench: Benchmarking Document-to-Database Extraction}
\title{Beyond Tables: Doc2DB-Bench for Relationally Faithful Document-to-Database Construction}

\makeatletter
\renewcommand{\@notice}{}
\makeatother
\author{
Zhuowen Liang$^{1}$,
Zhengxuan Zhang$^{1}$,
Jiayang Wang$^{1}$,
Jiazhuo Chen$^{1}$,
Nan Tang$^{1} \thanks{Corresponding author: Nan Tang (E-mail: nantang@hkust-gz.edu.cn)}$
 \vspace{.5em} 
  \\
$^{1}$The Hong Kong University of Science and Technology (Guangzhou) \\
}

\begin{document}

\maketitle

\begin{abstract}
Practical AI systems increasingly need to turn long, heterogeneous documents into queryable relational databases, not isolated spreadsheets. In domains such as finance, healthcare, education, transportation, and enterprise operations, downstream workflows rely on normalized schemas, entity identities, keys, cross-table relationships, and integrity constraints for analytics, compliance, auditing, and SQL-backed decision making. Existing Document-to-Table benchmarks are insufficient for this setting: flattening evidence into single tables can duplicate entities, obscure many-to-many relationships, create sparse records, and avoid testing whether extracted facts form a valid database instance. This creates an urgent need to evaluate document understanding as database construction rather than field extraction. We introduce \textbf{Doc2DB-Bench}, a benchmark for Document-to-Database construction, containing 203 long-document instances across 42 schemas and seven domain groups, with 117 entity tables, 132 relationship tables, 7,341 rows, and 41,935 cells. Built through a controllable DB-to-Doc synthesis pipeline and organized by a taxonomy of intra-table extraction and inter-table reasoning, the generated documents undergo authenticity verification, proving indistinguishable from real-world references. Doc2DB-Bench thus provides a testbed for reliable, auditable, and relationally faithful LLM-based data systems. The benchmark is
publicly available at \url{https://github.com/SetonLiang/Doc2DB-Bench}
\end{abstract}

\section{Introduction}
\label{sec:introduction}

Modern organizations rely on documents as the primary carrier of operational knowledge, from clinical notes and commercial contracts to financial reports and enterprise records~\cite{idc2023unstructured,chen2023symphony,li2026dataspace}. Yet downstream applications rarely consume free-form text directly: analytics pipelines, Business Intelligence dashboards, compliance workflows, and SQL-backed systems require structured, queryable, and auditable data~\cite{llamaindex2024extraction,zhang2025datamosaic,zeng2026qwenpaw}. As large language models become increasingly capable of processing long and heterogeneous documents, converting document evidence into reliable structured data has become a central goal of information extraction~\cite{xu2024llmie,liang2026long,li2025structrag}.

\textbf{Document-to-Table and Its Limitations.}
Most existing benchmarks study this problem under the \textit{Document-to-Table} (\doctable) setting, where systems extract fields or populate a single flattened table~\cite{dtbench2026,DBLP:conf/emnlp/DengC00FZYS24,DBLP:journals/corr/abs-2507-21340,datamosaic}. This setting is useful for isolated record extraction, but it is insufficient when downstream workflows require relational databases. Flattening multi-entity evidence can duplicate entities, obscure many-to-many relationships, introduce sparse records with excessive \texttt{NULL} values, and avoid testing whether extracted facts satisfy schema-level constraints.
\vspace{-0.5em}

\begin{figure}[t!]
    \centering
    \includegraphics[width=\textwidth]{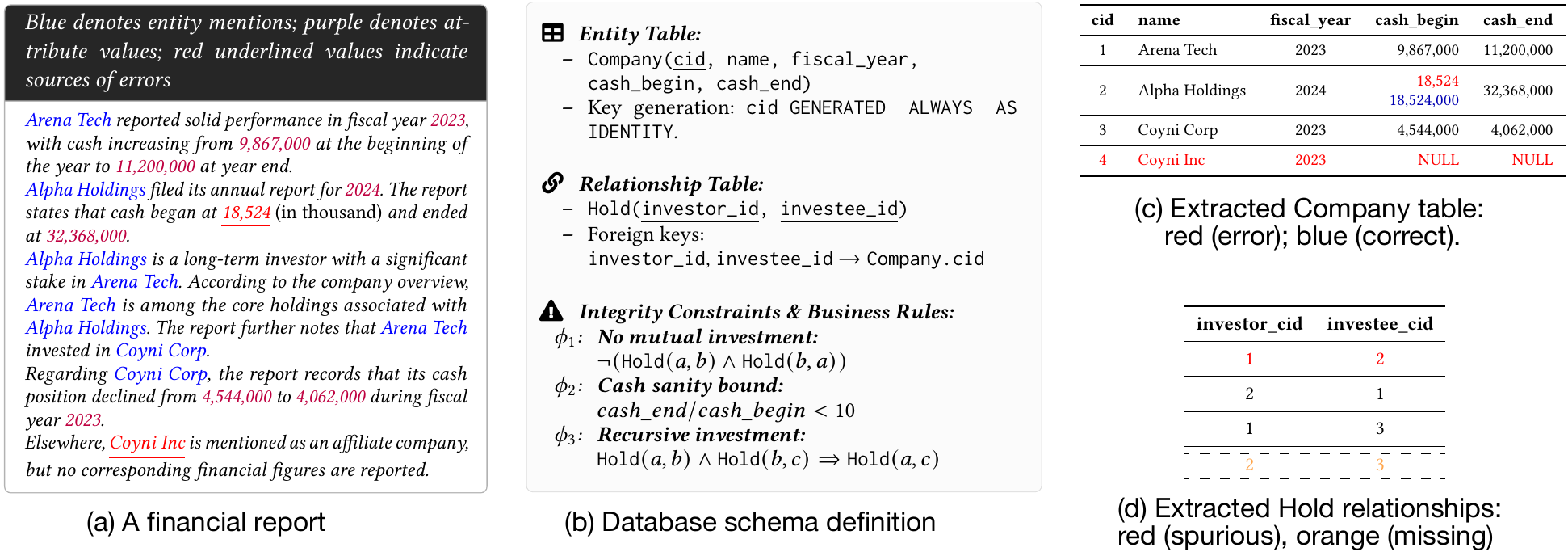}
    
    \caption{Document-to-Database extraction: (a) document(s); (b) target database schema; (c) Extracted Entity Tables; (d) Extracted Relationship Tables.}
    \vspace{-2em}
    \label{fig:mot}
\end{figure}

\begin{example}
\label{exam:extract}
Consider a financial report in Fig.~\ref{fig:mot}(a) describing multiple companies, their cash positions across fiscal years, and investment relationships among them. The target schema in Fig.~\ref{fig:mot}(b) contains an entity table \texttt{Company} and a self-referencing relationship table \texttt{Hold}, where company identifiers are system-generated keys and do not appear in the text.

Constructing the database requires more than extracting local values. An LLM extractor may correctly identify many company mentions and numbers, but still produce semantically invalid tuples. For entity extraction (Fig.~\ref{fig:mot}(c)), it may misinterpret units, such as treating ``18,524 (in thousand)'' as a raw value, or materialize weak mentions such as \emph{Coyni Inc} as incomplete duplicate entities. For relationship extraction (Fig.~\ref{fig:mot}(d)), surface-level cues may induce spurious mutual investments, while implicit multi-hop links may be missed. These errors show that local extraction decisions must be reconciled with schema-level semantics and integrity constraints.
\eop
\end{example}
\vspace{-1em}

\textbf{From Extraction to Database Construction.}
This motivates \textbf{Document-to-Database} (\docdb): given a document collection $F$, a target schema $E$, and integrity or business constraints $\Sigma$, the goal is to construct a relational database instance $D$ that is both faithful to the documents and valid under the schema:
$
(F,\; E,\; \Sigma)\;\xrightarrow{\;\textsc{Doc2DB}\;} D.
$
Unlike \doctable, \docdb is a database construction problem. Entity identifiers are often implicit, relationships may be distributed across document segments, and valid outputs must satisfy global constraints rather than independent field-level decisions~\cite{dong2014knowledge,sa2016deepdive,peng2017cross,shin2015incremental}. Therefore, evaluating document understanding at the database level requires testing not only value extraction, but also entity alignment, relationship construction, and relational validity.

Despite its practical importance, \docdb remains underexplored as a benchmark task. Existing information extraction benchmarks, including Rotowire~\cite{wiseman2017challenges}, E2E~\cite{DBLP:conf/sigdial/NovikovaDR17}, LiveSum~\cite{DBLP:conf/emnlp/DengC00FZYS24}, InstructIE~\cite{DBLP:conf/emnlp/Jiao0LZOJ023}, StructText~\cite{DBLP:journals/corr/abs-2507-21340}, and DTBench~\cite{dtbench2026}, mainly focus on flat tables, single-table extraction, or simplified generation tasks.
SQUiD~\cite{DBLP:conf/emnlp/SadiaYXCC25} explores text-to-relational database generation, but mainly targets logical relational view recovery rather than realistic \docdb construction.
As summarized in Table~\ref{tab:baselines}, they do not fully evaluate cross-table schema construction (i.e., normalized multi-table structures with inter-table dependencies), long-context evidence aggregation, and database-level correctness. 

\textbf{Challenges.}
Building a comprehensive \docdb benchmark is challenging for two reasons. First, direct human annotation is difficult to scale: annotators must collect documents, define schemas, and manually construct ground-truth entity and relationship tables. Second, realistic \docdb instances must cover diverse reasoning requirements, including unit normalization, ambiguity resolution, multi-hop relation construction, temporal changes, and constraint satisfaction, while also spanning domains with different narrative styles and schema structures.

\textbf{Our Proposal.}
We present \textbf{\ours}, a benchmark for evaluating \docdb capabilities beyond flat table extraction. To avoid the scalability bottleneck of manual annotation, we design a controllable \dbdoc reverse synthesis pipeline grounded in existing relational databases such as BIRD~\cite{li2023can} and Spider~\cite{yu2018spider}. Starting from schemas and database instances, the pipeline decomposes tuples into atomic evidence, assigns capability labels, serializes evidence into document plans, generates style-conditioned documents, and validates the generated documents through coverage and extraction-consistency checks.

To model realistic document complexity, \ours introduces a two-pillar \docdb taxonomy. \textit{Intra-Table Capabilities} cover cell-level extraction, normalization, inference, disambiguation, and faithfulness. \textit{Inter-Table Capabilities} capture database-specific reasoning, including identity resolution, relationship linking, multi-hop composition, dynamic change, and integrity constraints. This taxonomy supports controlled benchmark construction and fine-grained diagnosis of model failures. Finally, \ours evaluates outputs at the database level by measuring both entity integrity and relational fidelity, thereby testing whether extracted facts form a schema-conformant, queryable, and relationally faithful database instance.

\begin{table}[t]
\centering
\small
\label{tab:baselines}
\caption{Comparison of existing benchmarks and our \textsf{Doc2DB-Bench}.}
\vspace{-.5em}
\resizebox{\linewidth}{!}{%
\begin{tabular}{l|ccccc}
\toprule
Benchmark &
\makecell{Single-Table \\ Extraction (ST)} &
\makecell{Multi- \\ Domain (MD)} &
\makecell{Cross-Table \\ Reasoning (CR)} &
\makecell{Long- \\ Context (LC)} &
\makecell{DB-Level \\ Eval (DC)} \\
\midrule
Rotowire~\cite{wiseman2017challenges}   & \cmark & \xmark & \xmark & \xmark & \xmark \\
E2E~\cite{DBLP:conf/sigdial/NovikovaDR17}        & \cmark & \xmark & \xmark & \xmark & \xmark \\
LiveSum~\cite{DBLP:conf/emnlp/DengC00FZYS24}    & \cmark & \xmark & \xmark & \xmark & \xmark \\
InstructIE~\cite{DBLP:conf/emnlp/Jiao0LZOJ023} & \cmark & \cmark & \xmark & \xmark & \xmark \\
StructText~\cite{DBLP:journals/corr/abs-2507-21340} & \cmark & \cmark & \xmark & \xmark & \xmark \\
DTBench~\cite{dtbench2026}    & \cmark & \cmark & \xmark & \cmark & \xmark \\
SQUiD~\cite{DBLP:conf/emnlp/SadiaYXCC25} & \cmark & \cmark & \xmark & \xmark & \cmark \\

\midrule
\rowcolor{greyblue}
\texttt{\ours} (ours)
            & \cmark & \cmark & \cmark & \cmark & \cmark \\
\bottomrule
\end{tabular}%
}
\vspace{-1em}
\end{table}

\vspace{-.5em}

\stitle{Contributions.}
We summarize our contributions as follows:
\begin{enumerate}[leftmargin=*,itemsep=0.5pt,topsep=2pt]
    \item \textbf{Doc2DB Capability Taxonomy.}
    We introduce a two-pillar taxonomy of the key capabilities required for \docdb construction, spanning \textit{Intra-Table} value extraction and \textit{Inter-Table} relational reasoning, to guide benchmark design and support fine-grained evaluation.
    
    \item \textbf{Controllable \dbdoc Synthesis Pipeline.}
    We propose a reverse-synthesis pipeline that generates long-document \docdb instances from real relational schemas and database instances, reducing manual annotation cost while controlling document complexity.

    \item \textbf{The \ours Benchmark.}
    We construct a multi-domain benchmark for evaluating LLMs on relationally faithful \docdb tasks, going beyond flat, single-table extraction benchmarks.


    \item \textbf{Extensive Experiments.}
    We evaluate a broad range of mainstream LLMs on \ours and conduct in-depth analyses across entity- and relation-level capabilities, revealing their strengths, limitations, and open research opportunities for \docdb extraction.
\end{enumerate}
\vspace{-0.5em}

\vspace{-.5em}
\section{Document-to-Database: The Problem}
\label{sec:problem}

\subsection{Problem Statement}

\textbf{Database Specification.} 
We formalize the target database specification as
$\mathcal{S} = (E, \Sigma)$, where
$E = (\mathcal{E}, \mathcal{R})$ denotes the Entity-Relationship schema. $\mathcal{E}$ and $\mathcal{R}$ are the sets of entity and relationship tables, respectively, and
$\mathcal{T} = \mathcal{E} \cup \mathcal{R}$ denotes the set of tables. Each table $T \in \mathcal{T}$ is defined by an attribute set $A_T$ and a primary key
$K_T \subseteq A_T$.The constraint set
$\Sigma = \Sigma_{\mathrm{rel}} \cup \Sigma_{\mathrm{app}}$
consists of standard relational constraints, such as type, domain, and inclusion constraints,
and application-specific rules, such as temporal conditions.
Together, $\mathcal{S}$ specifies the target database structure and its validity conditions.

\textbf{Document Corpus.}
We consider a collection of heterogeneous documents
$F = \{f_1, f_2, \ldots, f_m\}$
that provide the evidence for instantiating the target schema.
We assume low-level processing, such as text extraction and layout analysis, has been completed.
Since real-world documents may contain incomplete information, unobserved attributes naturally remain
\texttt{NULL}.
Accordingly, every extracted value should be grounded in the source documents without hallucinating unsupported information.

\textit{Definition 1 (The \docdb Problem).}
Given a database specification $(E, \Sigma)$ and a document corpus $F = \{f_1, \ldots, f_m\}$, the \docdb task is to construct a relational database instance
$
(F, E, \Sigma) \xrightarrow{\;\docdb\;} D,
$
where $D = \{I_x \mid x \in E\}$ is the collection of instantiated entity and relationship tables. The target instance $D$ should satisfy (1) \textit{Constraint satisfaction}: $D \models \Sigma$; and (2) \textit{Closeness to ground truth}: $D$ is as close as possible to the ground-truth database $D^\ast$.

\vspace{-0.5em}
\subsection{\dbdoc Synthesis Paradigm}

To construct a benchmark for evaluating \docdb extraction, we adopt a reverse synthesis perspective: given a ground-truth database instance $D^{} = {I^{}_{x} \mid x \in E}$ with specification $\mathcal{S} = (E, \Sigma)$, we synthesize a source document corpus $F$ from which $D^{*}$ can be faithfully recovered.

\textit{Definition 2 (\dbdoc Synthesis).} 
Given a ground-truth database $D^\ast$, its specification $\mathcal{S}$, and a capability taxonomy $\mathcal{H}$,
\dbdoc synthesizes a document corpus $F$ satisfying three properties:
(1) \textit{completeness}, every ground-truth value and relational tuple in $D^\ast$ is supported by evidence in $F$;
(2) \textit{exclusiveness}, $F$ contains no evidence supporting database facts beyond $D^\ast$ under $\mathcal{S}$; and  (3) \textit{capability awareness}, the evidence for each target value/relational tuple is constructed to require specific reasoning capabilities from $\mathcal{H}$ for its recovery, with $\mathcal{H}$ formalized as a two-pillar taxonomy in Sec.~\ref{sec:doc2db_taxonomy}.

The completeness and exclusiveness conditions jointly ensure that $D^{*}$ is the unique ground-truth database recoverable from $F$ under $\mathcal{S}$, while capability awareness enables fine-grained evaluation across taxonomy levels. Each instance may exercise any subset of capabilities from $\mathcal{H}$, reflecting the naturally uneven distribution of requirements across real-world documents.

\vspace{-0.5em}
\section{\docdb Taxonomy}
\label{sec:doc2db_taxonomy}
\vspace{-0.5em}

\begin{figure*}[t!]
    \centering
    \includegraphics[width=\textwidth]{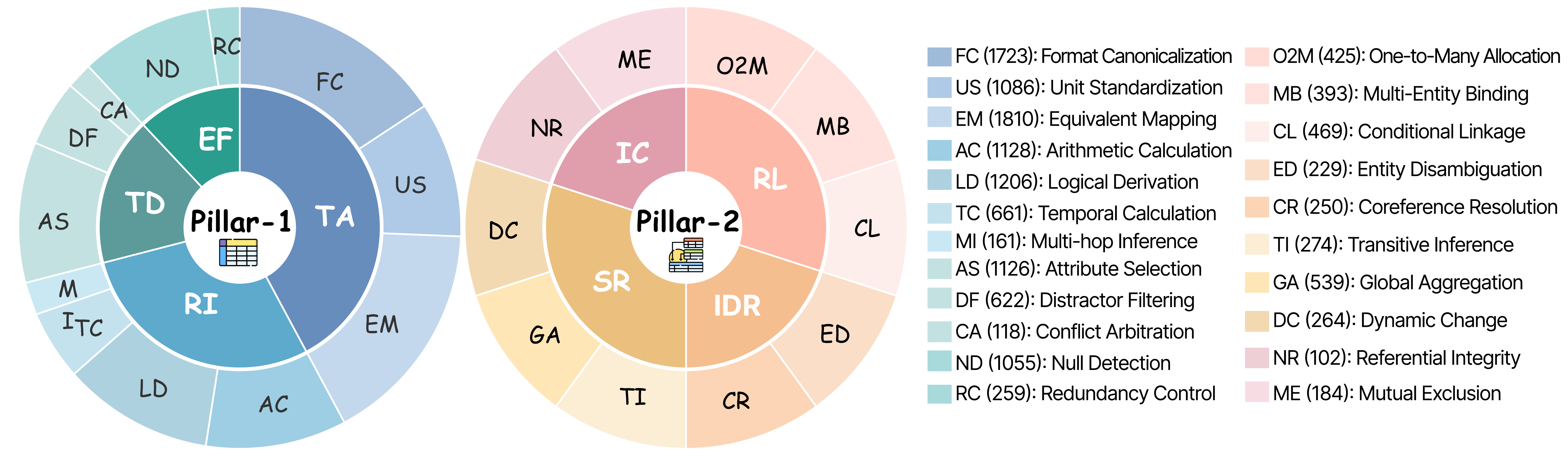}
    \caption{A Taxonomy of \docdb Extraction Capabilities.}
    \vspace{-1em}
    \label{fig:doc2db_taxonomy}
\end{figure*}

Doc2DB requires models to construct both values and relations. 
Accordingly, we organize the required capabilities into two complementary pillars. 
Intra-Table Capabilities (Pillar 1) asks \emph{what values should populate an entity record}, whereas 
Inter-Table (Pillar 2) asks \emph{how entity records should be connected}. 
Figure~\ref{fig:doc2db_taxonomy} summarizes the resulting taxonomy and its fine-grained sub-capabilities.

\vspace{-.5em}
\subsection{Pillar1: Intra-Table Capabilities}
\vspace{-0.5em}

Pillar 1 focuses on what values should populate an entity record, covering four complementary challenges.
\textbf{Transformative Alignment (TA).} Normalizes observed values into schema-compatible representations 
(\eg ``18,524 in thousands'' $\rightarrow 18{,}524{,}000$).
\textbf{Reasoning \& Inference (RI).} Derives values that are not explicitly stated 
(\eg computing a target value from several reported quantities).
\textbf{Target Discrimination (TD).} Identifies the intended value among distracting or conflicting evidence 
(\eg selecting the current-year rather than a historical figure).
\textbf{Evidence Faithfulness (EF).} Avoids unsupported outputs 
(\eg keeping an unreported attribute as \texttt{NULL}).
Pillar 1 extends DTBench~\cite{dtbench2026} from single-table extraction to schema-compliant entity construction in Doc2DB.

\vspace{-.5em}
\subsection{Pillar 2: Inter-Table Capabilities}
\vspace{-0.5em}

Pillar 2 focuses on how entity records should be connected, covering four complementary challenges.
\textbf{Identity Resolution (IDR).} Resolves ambiguous mentions to unique database entities 
(\eg identifying ``Lao Zhang'' as the employee in the technical department among records with the same name).
\textbf{Relationship Linking (RL).} Maps entity mentions and interactions into normalized relational tuples 
(\eg expanding ``Alice manages X, Y, and Z'' into three separate relationship rows).
\textbf{Structural Reasoning (SR).} Infers relations from distributed or compositional evidence 
(\eg ``Alice leads the Mobile Team'' and ``the Mobile Team manages Project X'' imply $(\text{Alice}, \text{Project X})$).
\textbf{Integrity Constraint (IC).} Suppresses relations that violate negative evidence or explicit constraints 
(\eg not linking Bob to Project X when the document states that all managers \emph{except Bob} are assigned to it).
Detailed sub-capabilities and examples are provided in Appendix~\ref{sec:app_taxonomy}.

\vspace{-0.5em}
\section{Document-to-Database (Db2Doc) Synthesis}
\label{sec:db2doc_pipeline}
\vspace{-0.5em}

\begin{figure*}[t!]
    \centering

    \includegraphics[width=1\textwidth]{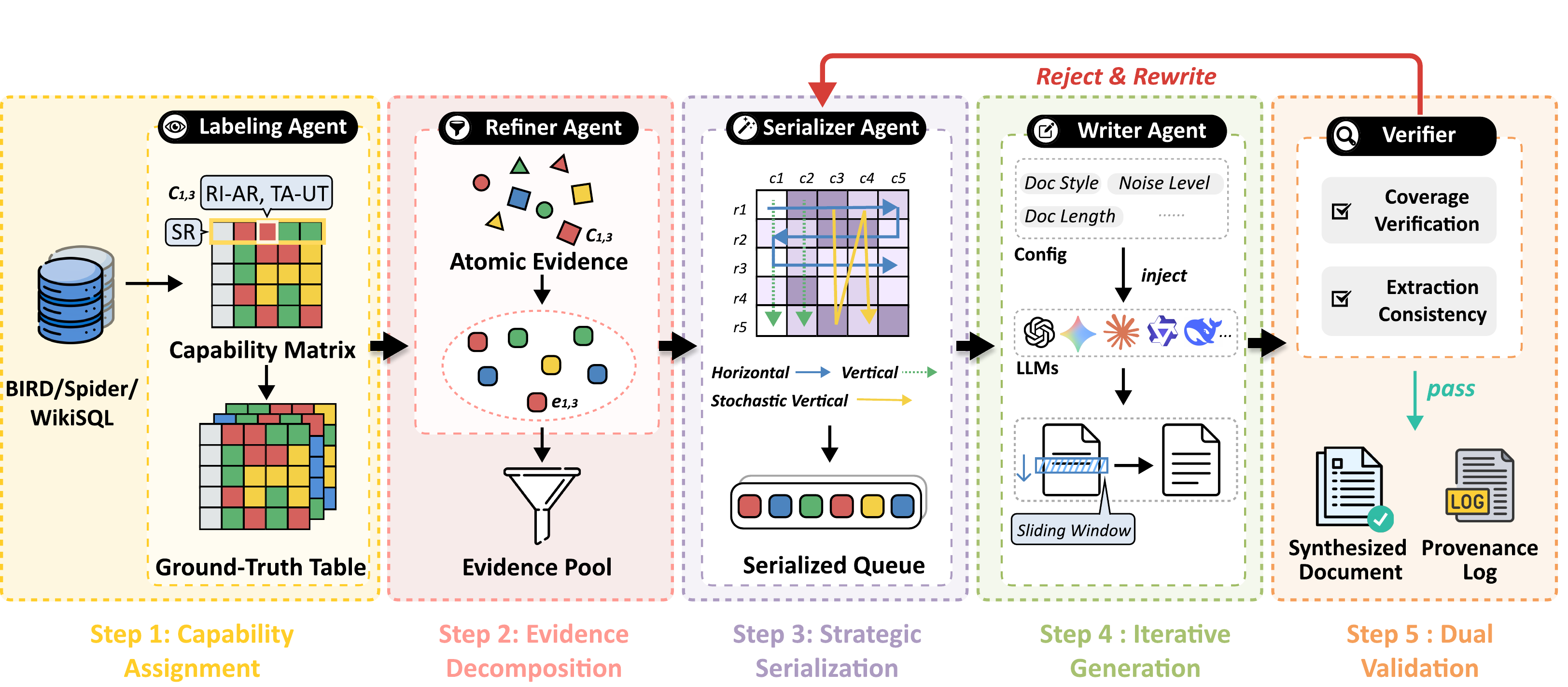}
    
    \caption{Overview of the benchmark construction framework.}
    \vspace{-2em}
    \label{fig:framework}
\end{figure*}

Given a ground-truth relational database instance $D$, its specification 
$S=(E,\Sigma)$, and the capability taxonomy $\mathcal{H}$, our goal is to 
synthesize a document corpus $F$ from which $D$ can be faithfully recovered. 
As shown in Fig.~\ref{fig:framework}, our DB2Doc pipeline consists of five 
stages: capability assignment, evidence decomposition, strategic serialization, 
iterative document generation, and dual validation.

The rationale behind this design is to separate \emph{what} should be tested 
from \emph{how} it is expressed in documents. Capability assignment determines 
the target reasoning requirements; evidence decomposition grounds each target 
value or tuple in atomic evidence; serialization controls the document structure; 
generation turns the evidence into realistic long-form text; and validation 
checks that the final document remains complete, faithful, and recoverable. 
This staged design makes the synthesis process controllable, reproducible, and 
suitable for fine-grained Doc2DB evaluation.

\textbf{Step 1: Capability Assignment.}
{\small\labelagent} A \underline{\textit{Labeling Agent}} scans each cell $v^{}_{ij} \in D^{}$ and assigns a label set $\mathcal{L}{ij} \subseteq \mathcal{H}$ to produce an annotated matrix \(\mathbf{M}_D\). The agent employs an LLM as a semantic router that jointly conditions on the cell value $v^{}{ij}$, its attribute context from $\mathcal{S}$, and the enclosing relational tuple $r$, reasoning over both intra-table semantics and inter-table relational structure to determine $\mathcal{L}{ij}$. Multi-label assignment captures compositional requirements, \eg a cell demanding summation followed by currency conversion is labeled $\mathcal{L}{ij} = {\texttt{RI-AR}, \texttt{TA-UT}}$, as illustrated in Sec.~\ref{sec:doc2db_taxonomy}. 

\textbf{Step 2: Joint Evidence Decomposition.}
{\small\refineragent}A \underline{\textit{Refiner Agent}} processes each annotated cell $C_{ij}$ and decomposes it into an atomic evidence set $E_{ij} = {e_1, e_2, \dots, e_k}$, forming the global evidence pool $\mathcal{P}$. Critically, the agent performs \textit{joint execution} over two orthogonal constraint dimensions simultaneously: the \textbf{attribute-level label} (Pillar 1, \eg \texttt{TA-EM} for enumeration mapping) and the \textbf{row-level label} (Pillar 2, \eg \texttt{SR} for structural reasoning).
To ensure evidence quality, we apply checklist-based verification along four dimensions:
(i) \textit{value/relation correctness}, ensuring factual consistency with target values and relations;
(ii) \textit{label faithfulness}, verifying that the assigned capability is properly instantiated;
(iii) \textit{schema linkage}, checking consistency with valid schema elements and cross-table references; and
(iv) \textit{data integrity and null faithfulness}, preventing unsupported content and preserving missing values.
Failed cases receive diagnostic feedback and are iteratively revised until verification succeeds or a retry limit is reached.
Detailed criteria are provided in Appendix~\ref{ssec:app_checklist}.

\textbf{Step 3: Strategic Serialization.} 
{\small\serializeragent} A \underline{\textit{Serializer Agent}} linearizes the verified evidence pool \(\mathcal{P}\) into coherent blocks, controlling the document's macro-structure and evidence organization.
Each evidence item \(e \in \mathcal{P}\) is indexed by its entity \(i\) and attribute \(j\), and serialized according to
\(V(e) = (\pi_1, \pi_2, \epsilon)\),
where \((\pi_1,\pi_2)\) denotes a mode-specific ordering of the entity and attribute indices \((i,j)\), and \(\epsilon \sim U(0,1)\) serves as a stochastic tie-breaker.
We consider three serialization strategies:
\textit{Horizontal} ($V=(i,j,\epsilon)$), which groups evidence by entity;
\textit{Vertical} ($V=(j,i,\epsilon)$), which groups evidence by attribute across entities; and
\textit{Stochastic Vertical} ($V=(\pi(j),i,\epsilon)$), which follows the vertical scheme while randomly permuting the attribute order.
Together, these strategies introduce controllable variation in document structure while maintaining coherent evidence organization.

\textbf{Step 4: Iterative Document Generation.}
{\small\generatoragent} A \underline{\textit{Writer Agent}} generates the final document block by block following the serialized plan. A context-aware sliding window conditions each block on the current evidence $E_{block}$ and a bounded history of previously generated text $H_{prev}$, promoting fluent transitions and consistent cross-block coreference.

Furthermore, to better emulate the complexity of real-world documents, generation is controlled by a \textbf{plug-and-play configuration module}.
It supports configurable parameters such as \texttt{Document\_Length}, \texttt{Noise\_Level} (\(\in [0,1]\)), and \texttt{Document\_Style}.
The noise level controls the frequency of natural distractors, including irrelevant background content, formatting artifacts, and header/footer interference, while the style parameter follows domain-specific reference documents (\eg formal legal writing or concise clinical shorthand).
Together, these controls increase document realism and diversity without violating the factual constraints in the evidence pool \(\mathcal{P}\).

\textbf{Step 5: Dual Validation.}
To guarantee that the synthesized document $D_{syn}$ satisfies the rigorous constraints of the \docdb paradigm, {\small\validatoragent} a \underline{\textit{Verifier Agent}} conducts a final dual-validation pass. This phase operates as an inverse-consistency check:
\begin{itemize}[leftmargin=*, nosep]
    \item \textbf{Coverage Verification:} Ensures that no required atomic evidence from \(\mathcal{P}\) is omitted or semantically distorted, utilizing explicit boundary tags (\eg \texttt{<frag\_}$i$\texttt{>}\dots\texttt{</frag\_}$i$\texttt{>}) injected during generation to deterministically trace textual spans back to their source atoms.
    \item \textbf{Extraction Consistency:} An evaluation mechanism is deployed to re-extract the database instance $D'_{syn}$ from $D_{syn}$. We rigorously verify the structural isomorphism and value equivalence between the re-extracted $D'_{syn}$ and the ground-truth database $D$.
\end{itemize}
If \(D_{\mathrm{syn}}\) fails either validation criterion due to hallucinated relations, missing evidence, or referential inconsistencies, it is routed back to Step 4 for targeted rewriting with error feedback.
This closed-loop validation minimizes unrecoverable or contradictory information in the final benchmark corpus.

\textbf{\ours Statistics.}
We synthesize documents from databases sourced from BIRD~\cite{li2023can} and Spider~\cite{yu2018spider}, using them as ground truth.
Specifically, we curate 42 high-quality databases across seven domains, including education and medical.
Details of database collection and processing are provided in Appendix~\ref{appendix:data_collection}.
All agents use Gemini-2.5-Pro~\cite{comanici2025gemini} as the backbone model, selected based on preliminary experiments balancing quality and cost.
All synthesized cases are further verified by seven computer science PhD candidates, each specializing in a corresponding domain.

As summarized in Table~\ref{tab:combined_stats}, \ours contains 42 synthesized (Document, Specification, Database) samples, covering 117 entity tables, 132 relationship tables, 7,341 rows, and 41,935 cells across seven domains.
The benchmark targets schema understanding, long-context reasoning, and cross-table dependency modeling, with fine-grained capability annotations for 11,205 cells (>25\%) and 3,129 rows (>40\%).
To guide realistic synthesis, we collect reference documents from DocBench~\cite{zou2025docbench}, MMLongBench~\cite{ma2024mmlongbench}, MMDocRAG~\cite{dong2025benchmarking}, and web sources, spanning financial and legal reports to research abstracts.
Fig.~\ref{fig:doc2db_taxonomy} shows the capability distribution; unannotated cells and rows correspond to directly extractable cases requiring verbatim recovery.

\begin{table*}[t]
\centering
\small
\caption{Comprehensive statistics of the \ours benchmark. The \textit{\#} denotes the total count, and \textit{Avg. Len.} indicates the average document length per sample in tokens.}
\label{tab:combined_stats}
\renewcommand{\arraystretch}{1.15}
\begin{tabular*}{\textwidth}{@{\extracolsep{\fill}} l cc ccc ccc @{}}
\toprule
\multirow{2}{*}{\textbf{Domain}} & \multicolumn{2}{c}{\textbf{Document}} & \multicolumn{3}{c}{\textbf{Schema}} & \multicolumn{3}{c}{\textbf{Table Scale (Total)}} \\
\cmidrule{2-3} \cmidrule{4-6} \cmidrule{7-9}
& \textbf{\# Docs} & \textbf{Avg. Len.} & \textbf{\# DBs} & \textbf{\# Ent.} & \textbf{\# Rel.} & \textbf{\# Rows} & \textbf{\# Cols} & \textbf{\# Cells} \\
\midrule
Education      & 13  & 40,424  & 3  & 12 & 12 & 830   & 96  & 3,451  \\
Finance        & 59  & 40,073  & 10 & 23 & 39 & 1,918 & 353 & 10,454 \\
Medical        & 11  & 43,930  & 3  & 9  & 12 & 343   & 199 & 3,522  \\
University     & 31  & 33,913  & 6  & 17 & 21 & 860   & 209 & 4,339  \\
Sports         & 27  & 50,991  & 5  & 20 & 16 & 1,085 & 177 & 5,772  \\
Transportation & 17  & 53,091  & 5  & 12 & 6  & 718   & 128 & 4,678  \\
Others         & 43  & 46,566  & 10 & 24 & 26 & 1,587 & 313 & 9,719  \\
\midrule
\textbf{Total/Avg.} & \textbf{203} & \textbf{43,326} & \textbf{42} & \textbf{117} & \textbf{132} & \textbf{7,341} & \textbf{1,475} & \textbf{41,935} \\
\bottomrule
\end{tabular*}
\vspace{-2em}
\end{table*}

\vspace{-1em}
\section{Experiments}
\label{sec:experiments}
\vspace{-0.5em}

In this section, we systematically evaluate the performance of various LLMs on \ours for the \docdb extraction task. Specifically, we seek to answer the following research questions:

\noindent\textbf{RQ1:}  How well do different LLMs perform on \docdb extraction at the entity and relation levels?


\noindent\textbf{RQ2:} Beyond overall performance, what specific capabilities and vulnerabilities do these models exhibit across the fine-grained dimensions defined in our taxonomy?

\noindent\textbf{RQ3:}  How realistic and authentic are the documents generated by our automated pipeline?


\begin{table}[t]
  \centering
  \small
  \setlength{\tabcolsep}{6.5pt}
  \renewcommand{\arraystretch}{1.35}
  \caption{Comparative performance of LLMs on \ours. \colorbox{firstBest}{\textbf{Green}} highlights the best.}
  \vspace{-1em}
  \resizebox{\textwidth}{!}{
  \begin{tabular}{l c|ccc|ccc|cccc}
      \toprule
      \multirow{2}{*}{\textbf{Model}} & \multirow{2}{*}{\textbf{Size}} 
      & \multicolumn{3}{c|}{\textbf{Entity-Level}}
      & \multicolumn{3}{c|}{\textbf{Relation-Level}} 
      & \multicolumn{4}{c}{\textbf{Overall Performance}} \\
      
      & & P. & R. & F1
        & P. & R. & F1
        & P. & R. & F1 & \textcolor{blue}{LS} \\
        
      \midrule

      \rowcolor{gray!10}
      \multicolumn{12}{c}{\textit{Open-source Models}} \\
      \midrule
              
      
      Qwen2.5-14b-Ins & 14b & 42.87	& 60.17 &	44.07 & 27.45	& 43.54	& 28.92 & 29.40 & 50.36	& 32.97 & \textcolor{blue}{39.39} \\
      
      
      
      LLaMA-3.1-70b-Ins & 70b & 14.96	& 58.89	& 17.25 & 20.29 & 38.79	& 16.97 & 12.12 & 47.46 & 14.29 & \textcolor{blue}{22.76} \\
      
      Qwen2.5-72b-Ins & 72b & 41.62 & 66.57 &	45.03 & 28.48	& 54.63	& 31.98	 & 30.96	& 59.53 & 36.07 & \textcolor{blue}{39.53} \\

      \midrule

      \rowcolor{gray!10}
      \multicolumn{12}{c}{\textit{Proprietary Models}} \\
      \midrule
      
      GPT-4o~\cite{hurst2024gpt} & - & 72.39 & 70.92 & 70.30 & 62.94 & 46.78 & 50.68 & 65.66 & 55.99 & 59.05 & \textcolor{blue}{47.52} \\

      
      GPT-5.4 & - & 79.93 & 81.12 & 80.09 & \colorbox{firstBest}{80.09} & \colorbox{firstBest}{69.63} & \colorbox{firstBest}{72.33} & \colorbox{firstBest}{78.47} & \colorbox{firstBest}{73.93} & \colorbox{firstBest}{75.25} & \colorbox{firstBest}{\textcolor{blue}{69.02}} \\

      Gemini-2.5-flash~\cite{comanici2025gemini} & - & 75.89 & 75.48 & 74.21 & 73.60 & 57.35 & 59.50 & 71.33 & 64.58 &	65.99 & \textcolor{blue}{53.74} \\
      
      Gemini-2.5-pro~\cite{comanici2025gemini} & - & 76.80 & 82.24 & 78.13 & 67.98 & 67.01 & 65.46 & 69.83 & 73.80 & 70.81 & \textcolor{blue}{57.19} \\ 
              
      
      Claude-opus-4-6 & - & \colorbox{firstBest}{87.19} & \colorbox{firstBest}{83.70} & \colorbox{firstBest}{84.93} & 66.71 & 62.16 & 63.17 & 77.33 & 71.57 & 73.60 & \textcolor{blue}{61.95} \\ 

      Qwen3-max & - & 76.51 & 77.00& 75.77 & 57.01 & 59.74 & 55.96 & 63.48 & 66.42 & 63.44 & \textcolor{blue}{52.48} \\
      
      Deepseek-V4-flash & - & 76.65 &66.06 &	67.30 & 67.67 & 43.23& 48.93 & 69.03 &	52.14	& 57.06 & \textcolor{blue}{49.05} \\
      
      \midrule

      \rowcolor{gray!10}
      \multicolumn{12}{c}{\textit{Specialized IE Systems}} \\
      \midrule

      LangExtract~\cite{langextract2025google} & - & 29.72 & 40.50 & 33.49 & 33.34 & 19.08 & 22.00 & 36.06 & 25.20 & 28.72 & \textcolor{blue}{23.55} \\

      DocETL~\cite{shankar2024docetl} & - & 78.20 & 76.85 & 74.91 & 46.75 & 41.26 & 41.51 & 65.27 & 60.97 & 60.82 & \textcolor{blue}{49.87} \\

      LangChain~\cite{langchain2022} & - & 60.77 & 83.09 & 69.09 & 33.50 & 42.47 & 36.33 & 47.28 & 64.38 & 53.83 & \textcolor{blue}{44.14} \\

      LlamaExtract~\cite{llamaindex2024extraction} & - & 52.28 & 68.95 & 56.58 & 47.41 & 48.73 & 47.39 & 45.35 & 56.49 & 49.10 & \textcolor{blue}{40.26} \\
      
      \bottomrule
  \end{tabular}}
  \label{tab:main_results_1}
\vspace{-1em}

\end{table}

\vspace{-.5em}
\subsection{Setup}
\label{ssec: setup}
\vspace{-.5em}

\textbf{Baselines.}
We benchmark a broad range of state-of-the-art LLMs, categorized into three groups:
(1) \textit{Open-weight models}, including Qwen2.5-14B/72B-Instruct~\cite{DBLP:journals/corr/abs-2412-15115} and Llama-3.1-70B-Instruct;
(2) \textit{Proprietary APIs}, including GPT-4o~\cite{hurst2024gpt}, GPT-5.4, Gemini-2.5-Flash/Pro~\cite{comanici2025gemini}, Claude-opus-4-6~\cite{anthropic2026claudeopus46}, Qwen3-Max, and DeepSeek-V4-Flash~\cite{xu2026deepseek}; and
(3) \textit{Specialized IE systems}, including LlamaExtract~\cite{llamaindex2024extraction}, a commercial extraction service with iterative refinement and citations; LangExtract~\cite{langextract2025google}, which uses chunking and few-shot prompting for long-document extraction; LangChain~\cite{langchain2022}, a general framework with schema-constrained function calling for structured outputs; and DocETL~\cite{shankar2024docetl}, a declarative LLM-based data processing framework with modular extraction pipelines. 


\noindent\textbf{Evaluation Details.}
For fair and reproducible comparison, we use identical prompts and greedy decoding with temperature $0$ across all experiments.
All systems use GPT-5.4 as the backbone model, except LlamaExtract, which uses its proprietary model.
To isolate relational reasoning from upstream entity extraction errors, we additionally introduce an \textbf{oracle entity setting}, where the model $\mathcal{M}$ receives ground-truth entity tables $E^*$ and predicts only relation tables:
$T_{\text{rel}}=\mathcal{M}(D,S,E^*)$.

\noindent\textbf{Metrics.} To evaluate extraction at a granular level, we
employ the following metrics:
\vspace{-.5em}

\begin{itemize}[leftmargin=*, topsep=2pt, itemsep=2pt]

\item \textbf{Cell-level Accuracy.}
We evaluate fine-grained attribute extraction using cell-level Precision, Recall, and F$_1$. 
Rather than relying on row-level matching, which is sensitive to tuple ordering and may cause cascading alignment errors, we employ a \textbf{Global Maximum Weight Matching} algorithm to establish one-to-one tuple alignment between predicted ($T$) and ground-truth ($T^*$) tables (details in Appendix~\ref{ssec:app_metric}).
Let $C_{\text{match}}$ denote the number of cells satisfying our matching criteria (\eg, exact numeric equality or ${\geq}90\%$ string similarity). The metrics are defined as:
\[
P = \frac{|C_{\text{match}}|}{C_{\text{total}}(T)}, \quad
R = \frac{|C_{\text{match}}|}{C_{\text{total}}(T^*)}, \quad
F_1 = \frac{2PR}{P+R},
\]
where $C_{\text{total}}(\cdot)$ denotes the number of non-empty cells in the corresponding table.

\item \textbf{Semantic Quality.}
We employ GPT-4o as an evaluator to assess semantic equivalence beyond exact cell matching. 
It assigns a score from 0 to 100 based on factual correctness, schema conformity, and robustness to paraphrasing, capturing semantically equivalent but structurally varied predictions. 
The evaluation prompt is provided in Appendix~\ref{ssec: app_llmscore}, with further results in Appendix~\ref{sec:appendix_further_result}.


\end{itemize}

\vspace{-.5em}
\subsection{Experimental Results and Analysis}
\label{ssec:overall_performance}

\textbf{Overall Performance. }
As shown in Table~\ref{tab:main_results_1}, proprietary models consistently outperform open-source counterparts across all metrics. In particular, GPT-5.4 achieves the best overall F1 score (75.25) and the highest llm score (69.02), indicating strong capability in structured extraction.  Claude-opus-4-6 and Gemini-2.5-pro also deliver competitive performance. Notably, GPT-4o shows only moderate results (Overall F1 59.05), falling noticeably short of reasoning-capable models despite its scale. In contrast, open-source models such as Qwen2.5-14B-Ins exhibit significantly lower overall F1 (32.97), highlighting the persistent performance gap under complex extraction settings.

\vspace{-0.5em}
\begin{findingbox}[attach title to upper,after title={.\ }]{Finding 1}
Models with stronger reasoning capabilities consistently achieve better performance on Doc2DB extraction, where structural reasoning proves critical for schema-compliant output.
\end{findingbox}
\vspace{-0.5em}

\textbf{Entity-level performance.}
Entity-level results reveal a clear advantage of high-capacity proprietary models. Claude-opus-4-6 achieves the highest entity-level F1 (84.95), followed by GPT-5.4 (80.09) and Gemini-2.5-pro (78.13). These models maintain a good balance between precision and recall, indicating robust span detection and boundary alignment. Compared to them, Qwen2.5-14B shows notably lower precision (42.87) despite relatively higher recall (60.17), suggesting that open-source models tend to over-generate entity spans, leading to reduced accuracy.

\textbf{Relationship level performance.}
Relation extraction is evaluated under an oracle entity setting: given the relational schema, the source document, and the ground-truth entity tables, the model generates only the relation tables. This setting isolates relational reasoning ability from upstream entity extraction errors. As shown in Table~\ref{tab:main_results_1}, relation-level extraction remains substantially more challenging than entity extraction for most models. A key reason is that relation prediction requires not only identifying the correct records, but also globally aligning related entities and tuples under a structured schema, which is more error-prone than localized entity extraction. GPT-5.4 achieves the best relation-level F1 (72.33), outperforming GPT-4o (50.68), Gemini-2.5-pro (65.46), and Claude-opus-4-6 (63.13), indicating stronger capability in global alignment and structured relational reasoning.

\vspace{-0.5em}
\begin{findingbox}[attach title to upper,after title={.\ }]{Finding 2}
Relation-level extraction is a major bottleneck in Doc2DB extraction. Compared with entity extraction, it is substantially more challenging because it requires accurate global alignment across entities and tuples, leading to consistently lower performance.
\end{findingbox}
\vspace{-0.5em}


\begin{table*}[t!]
\renewcommand{\arraystretch}{1.6}
\centering
\caption{The Comparison of Capabilities Across Different Models for Different Categories. Cell colors represent performance levels (Blue: High; Red: Low).}
\resizebox{\textwidth}{!}{
\begin{tabular}{|c|c|c|c|c|c|c|c|c|c|}
\hline
\textbf{Capability} &
  \textbf{Sub-capabilities} &
  \multicolumn{1}{c|}{\textbf{\begin{tabular}[c]{@{}c@{}}Qwen2.5-\\14B-Ins\end{tabular}}} &
  \multicolumn{1}{c|}{\textbf{\begin{tabular}[c]{@{}c@{}}LLaMA-3.1-\\70B-Ins\end{tabular}}} &
  \multicolumn{1}{c|}{\textbf{GPT-4o}} &
  \multicolumn{1}{c|}{\textbf{\begin{tabular}[c]{@{}c@{}}DeepSeek-\\V4-Flash\end{tabular}}} &
  \multicolumn{1}{c|}{\textbf{Qwen3-Max}} &
  \multicolumn{1}{c|}{\textbf{\begin{tabular}[c]{@{}c@{}}Claude-\\Opus-4-6\end{tabular}}} &
  \multicolumn{1}{c|}{\textbf{\begin{tabular}[c]{@{}c@{}}Gemini-\\2.5-Pro\end{tabular}}} &
  \multicolumn{1}{c|}{\textbf{GPT-5.4}} \\ \hline

&
  \textbf{Entity Disambiguation} &
  \cellcolor[HTML]{F9FCFE}51.58 &
  \cellcolor[HTML]{D7EBF8}59.92 &
  \cellcolor[HTML]{B1D8F1}69.24 &
  \cellcolor[HTML]{EBF5FC}55.25 &
  \cellcolor[HTML]{C4E1F5}64.57 &
  \cellcolor[HTML]{A2D0EF}72.96 &
  \cellcolor[HTML]{86C2EA}80.03 &
  \cellcolor[HTML]{88C3EA}79.49 \\ \cline{2-10}
\multirow{-2}{*}{\textbf{\begin{tabular}[c]{@{}c@{}}Identity \\ Resolution\end{tabular}}} &
  \textbf{Coreference Resolution} &
  \cellcolor[HTML]{FCE5E3}37.84 &
  \cellcolor[HTML]{FFFFFF}50.13 &
  \cellcolor[HTML]{D2E9F7}61.14 &
  \cellcolor[HTML]{CEE7F7}62.10 &
  \cellcolor[HTML]{E4F2FA}56.76 &
  \cellcolor[HTML]{B8DBF3}67.61 &
  \cellcolor[HTML]{8FC7EB}77.71 &
  \cellcolor[HTML]{7DBDE8}82.18 \\ \hline

&
  \textbf{One-to-Many Allocation} &
  \cellcolor[HTML]{FFFEFD}49.06 &
  \cellcolor[HTML]{F8FCFE}51.90 &
  \cellcolor[HTML]{C0DFF4}65.63 &
  \cellcolor[HTML]{CDE6F6}62.27 &
  \cellcolor[HTML]{C4E1F5}64.77 &
  \cellcolor[HTML]{ADD6F1}70.38 &
  \cellcolor[HTML]{8DC6EB}78.18 &
  \cellcolor[HTML]{7FBEE8}81.83 \\ \cline{2-10}
&
  \textbf{Multi-Entity Binding} &
  \cellcolor[HTML]{FFFAFA}47.65 &
  \cellcolor[HTML]{F0F8FD}53.85 &
  \cellcolor[HTML]{D6EBF8}60.18 &
  \cellcolor[HTML]{F0F8FD}53.83 &
  \cellcolor[HTML]{DDEEF9}58.57 &
  \cellcolor[HTML]{C3E1F5}64.82 &
  \cellcolor[HTML]{BADCF3}67.26 &
  \cellcolor[HTML]{7ABCE8}82.97 \\ \cline{2-10}
\multirow{-3}{*}{\textbf{\begin{tabular}[c]{@{}c@{}}Relationship \\ Linking\end{tabular}}} &
  \textbf{Conditional Linkage} &
  \cellcolor[HTML]{F7FBFE}52.02 &
  \cellcolor[HTML]{E1F0FA}57.47 &
  \cellcolor[HTML]{B2D8F2}69.10 &
  \cellcolor[HTML]{CCE5F6}62.65 &
  \cellcolor[HTML]{D0E7F7}61.80 &
  \cellcolor[HTML]{B0D7F1}69.64 &
  \cellcolor[HTML]{97CBED}75.81 &
  \cellcolor[HTML]{89C4EA}79.16 \\ \hline

&
  \textbf{Transitive Inference} &
  \cellcolor[HTML]{FDECEA}40.80 &
  \cellcolor[HTML]{E0EFFA}57.83 &
  \cellcolor[HTML]{C1E0F4}65.51 &
  \cellcolor[HTML]{E2F1FA}57.35 &
  \cellcolor[HTML]{F0F8FD}53.78 &
  \cellcolor[HTML]{D9ECF9}59.48 &
  \cellcolor[HTML]{ABD5F0}70.82 &
  \cellcolor[HTML]{8CC5EB}78.47 \\ \cline{2-10}
&
  \textbf{Global Aggregation} &
  \cellcolor[HTML]{FFFEFE}49.23 &
  \cellcolor[HTML]{FFFCFC}48.30 &
  \cellcolor[HTML]{A8D3F0}71.59 &
  \cellcolor[HTML]{D8ECF9}59.83 &
  \cellcolor[HTML]{CEE7F7}62.15 &
  \cellcolor[HTML]{C6E2F5}64.25 &
  \cellcolor[HTML]{A2D0EF}73.03 &
  \cellcolor[HTML]{7DBDE8}82.20 \\ \cline{2-10}
\multirow{-3}{*}{\textbf{\begin{tabular}[c]{@{}c@{}}Structural \\ Reasoning\end{tabular}}} &
  \textbf{Dynamic Change} &
  \cellcolor[HTML]{FCE3E0}36.51 &
  \cellcolor[HTML]{F8FCFE}51.83 &
  \cellcolor[HTML]{DFEFFA}58.11 &
  \cellcolor[HTML]{FFF4F4}46.93 &
  \cellcolor[HTML]{DEEEF9}58.36 &
  \cellcolor[HTML]{BEDEF4}66.26 &
  \cellcolor[HTML]{8BC5EB}78.69 &
  \cellcolor[HTML]{89C4EA}79.27 \\ \hline

&
  \textbf{Referential Integrity} &
  \cellcolor[HTML]{FFFFFF}- &
  \cellcolor[HTML]{FFFFFF}- &
  \cellcolor[HTML]{FFFFFF}- &
  \cellcolor[HTML]{FFFFFF}- &
  \cellcolor[HTML]{FFFFFF}- &
  \cellcolor[HTML]{FFFFFF}- &
  \cellcolor[HTML]{FFFFFF}- &
  \cellcolor[HTML]{FFFFFF}- \\ \cline{2-10}
\multirow{-2}{*}{\textbf{\begin{tabular}[c]{@{}c@{}}Integrity \\ Constraint\end{tabular}}} &
  \textbf{Mutual Exclusion} &
  \cellcolor[HTML]{FCE2E0}36.41 &
  \cellcolor[HTML]{E3F1FA}56.94 &
  \cellcolor[HTML]{D6EBF8}60.18 &
  \cellcolor[HTML]{DDEEF9}58.45 &
  \cellcolor[HTML]{EDF6FC}54.61 &
  \cellcolor[HTML]{CFE7F7}61.96 &
  \cellcolor[HTML]{ADD6F1}70.39 &
  \cellcolor[HTML]{9DCDEE}74.46 \\ \hline

\end{tabular}
}
\label{table:capability_subtype}
\end{table*}




    

\begin{figure*}[t!]
    \vspace{-1em}
    \centering
    \begin{minipage}[b]{0.48\textwidth}
        \centering
        \includegraphics[width=\textwidth]{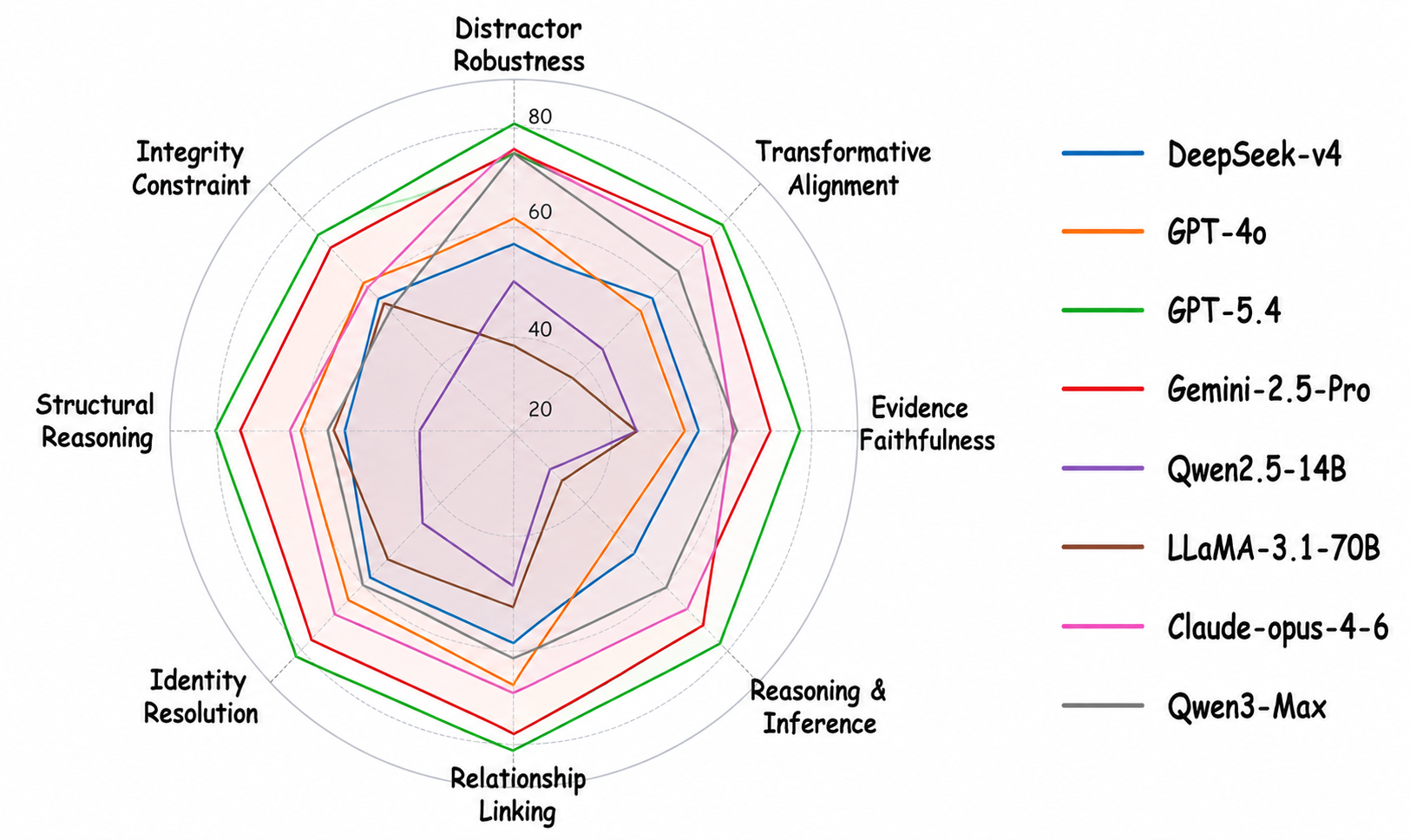}
        \caption{Radar chart of different models across
        various capabilities.}
        \label{fig:taxonomy_radar}
    \end{minipage}
    \hfill 
    \begin{minipage}[b]{0.48\textwidth}
        \centering
        \includegraphics[width=\textwidth]{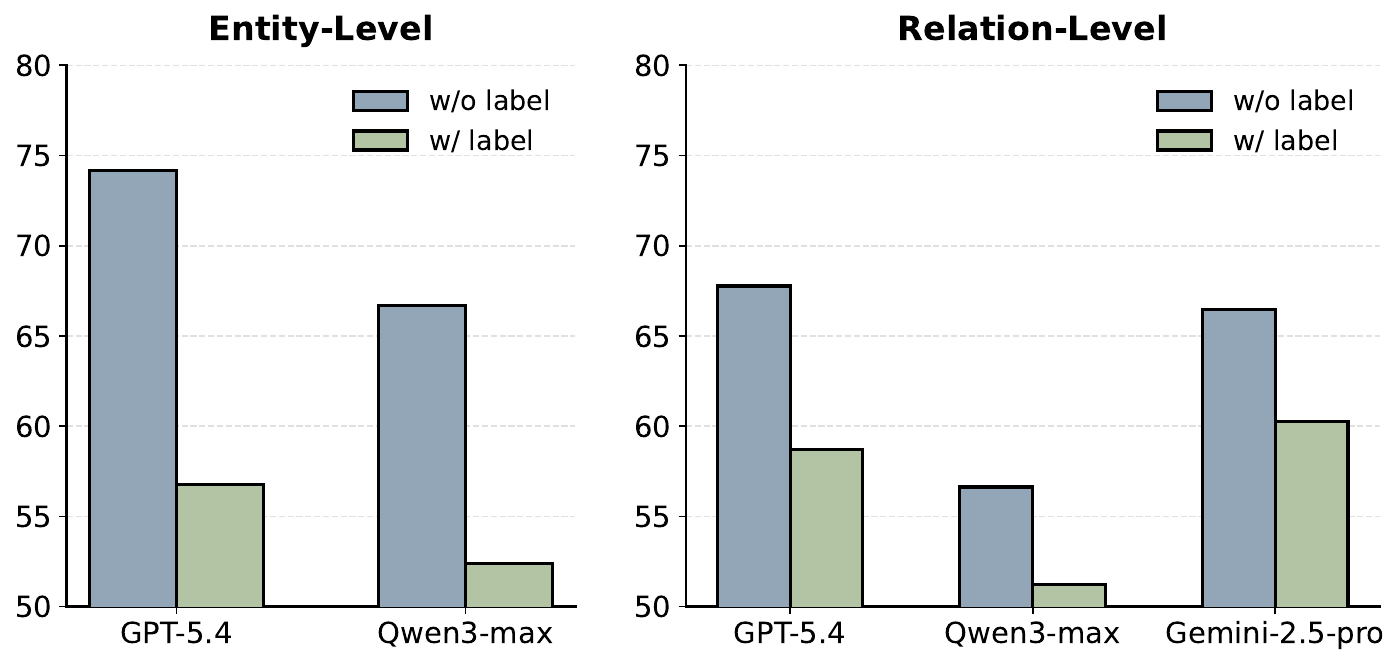}
        \caption{Impact of capability annotations on extraction performance.}
        \label{fig:label_ablation}
    \end{minipage}
    \vspace{-2em}
\end{figure*}

\subsection{Capability-Level Performance}

As revealed in Section~\ref{ssec:overall_performance}, relation extraction is the primary bottleneck in Doc2DB extraction, yet overall metrics do not reveal \textit{which specific inter-table capabilities} remain challenging for LLMs. We therefore conduct a fine-grained evaluation over Pillar~2 capabilities in our taxonomy.

For each capability, we compute cell-level Precision, Recall, and F$_1$ on the corresponding annotated relation tables using the same matching criteria, enabling fine-grained analysis of inter-table reasoning abilities such as entity linking, relation alignment, and multi-hop dependency recovery.

\noindent\textbf{Capability assignment introduces difficulty.}
As shown in Fig.~\ref{fig:label_ablation}, models perform consistently worse on documents with explicit capability labels than on those without. The drop is substantial at the entity level, with GPT-5.4 decreasing from 74.15 to 56.75 and Qwen3-max from 66.70 to 52.37. At the relation level, the decline is smaller but consistent across models (GPT-5.4: 67.77 to 58.69; Qwen3-max: 56.63 to 51.23; Gemini-2.5-pro: 66.49 to 60.26). These results confirm that capability-annotated documents are genuinely harder and validate the effectiveness of our taxonomy.

\noindent\textbf{Detailed Error Analysis.}
Fig.~\ref{fig:taxonomy_radar} shows that GPT-5.4 performs best across all capabilities, while Table~\ref{table:capability_subtype} reports fine-grained results across four inter-table capabilities, revealing three key findings: \noindent \textit{(i) Implicit evidence amplifies model disparities.} Surface-level tasks rely on explicit cues, whereas coreference resolution and multi-entity binding require aggregating scattered implicit evidence, significantly widening the gap between strong and weak models. \noindent \textit{(ii) Structural reasoning remains a major bottleneck.} Even SOTA models plateau at around 80\% on Dynamic Change and Transitive Inference, while smaller models fall below 40\%, indicating limited multi-step relational reasoning ability. \noindent \textit{(iii) Integrity Constraint (IC) verification is consistently weak.} Models often hallucinate dependencies in Referential Integrity tests and violate exclusivity rules in Mutual Exclusion tasks, highlighting the lack of constraint-aware generation and potential risks to database reliability.

\vspace{-.5em}
\begin{findingbox}[attach title to upper,after title={.\ }]{Finding 3.}
Implicit reasoning over distributed and compositional evidence increases difficulty and widens model gaps, with structural reasoning and integrity constraints as the primary failure modes.
\end{findingbox}
\vspace{-.5em}

\begin{table}[t]
\centering
\caption{Quality and Authenticity evaluation of \ours documents using (a) rubric-based LLM and human assessment and (b) commercial AI detectors. }
\label{tab:authenticity}
\vspace{-0.5em}

    
\begin{minipage}[t]{0.56\textwidth}
    \centering
    \small
    \setlength{\tabcolsep}{4pt}
    \renewcommand{\arraystretch}{1.3}

    \resizebox{\linewidth}{!}{
    \begin{tabular}{cccccc}
    \toprule
    \multirow{2}{*}{\textbf{Judger}}
    & \textbf{Lexical}
    & \textbf{Logical}
    & \textbf{Textual}
    & \textbf{Genre}
    & \multirow{2}{*}{\textbf{Average}} \\
    
    & \textbf{Richness}
    & \textbf{Consistency}
    & \textbf{Coherence}
    & \textbf{Fidelity}
    & \\
    \midrule

    LLM   & 4.46 & 4.28 & 3.84 & 4.94 & 4.38 \\
    Human & 4.17 & 3.97 & 3.98 & 4.00 & 4.03 \\
    \bottomrule
    \end{tabular}
    }
\end{minipage}%
\hfill
\begin{minipage}[t]{0.42\textwidth}
    \centering
    \small
    \setlength{\tabcolsep}{4pt}
    \renewcommand{\arraystretch}{1.1}

    \resizebox{\linewidth}{!}{%
    \begin{tabular}{lcc}
    \toprule
    \textbf{Method}
    & \textbf{Doc2DB-Bench}
    & \textbf{Reference} \\
    \midrule
    ZeroGPT & 35.20 & 23.12 \\
    FastGPT~\cite{bao2023fast} & 37.50 & 35.38 \\
    \bottomrule
    \end{tabular}%
    }
\end{minipage}
\vspace{-1em}
\end{table}

\subsection{Authenticity Verification}
\vspace{-0.5em}


To evaluate document quality and realism, we conduct a rubric-based assessment using both an LLM judge and human annotators.
Documents are evaluated along four dimensions: \emph{lexical richness, logical consistency, textual coherence, and genre fidelity}, using a five-point scale.
We use DeepSeek-V3.2 as the automatic evaluator and recruit five graduate students document intelligence research experience to independently assess 50 randomly sampled documents using the same rubrics.
As shown in Table~\ref{tab:authenticity}(a), the LLM and human evaluations consistently demonstrate the high linguistic quality and domain authenticity.
Furthermore, commercial AI detectors (ZeroGPT and FastGPT~\cite{bao2023fast}) show that \ours closely matches real-world distributions in AI-generation scores (Table~\ref{tab:authenticity}(b)). Notably, the FastGPT score (37.50\%) is nearly identical to the real-data baseline (35.38\%).
Overall, these results demonstrate that \ours preserves the lexical diversity and stylistic nuances of human-written reports, achieving high fidelity. More details are provided in Appendix~\ref{sec: appendix_authenticity}.

\section{Related Work}
\label{sec:related_work}
\vspace{-.5em}

\textbf{Structured Information Extraction.}
Traditional IE decomposes the problem into sub-tasks such as named entity
recognition and relation extraction, evolving from rule-based
methods~\cite{hobbs1993fastus,lee2013attribute} to pretrained language
model-based approaches~\cite{li2022sparse,wu2008information,zheng2018opentag}.
Such pipelines are difficult to customize for schema-driven extraction, where
target attributes may fall outside existing ontologies~\cite{DBLP:conf/emnlp/Jiao0LZOJ023}.
Recent work leverages LLMs for structured extraction: some advanced prompting methods ~\cite{DBLP:conf/emnlp/DengC00FZYS24,ahuja2025map,tang2024struc} decompose complex extraction tasks into intermediate steps; StructSum~\cite{jain2024structsum}
generates structured summaries as tables and mind maps; InstructIE~\cite{DBLP:conf/emnlp/Jiao0LZOJ023},
Lotus~\cite{patel2025semantic}, and PZ~\cite{liu2025palimpzest} extract specified attributes through natural-language queries; Doctopus~\cite{chai2025doctopus}
combines LLMs with traditional IE tools to reduce cost. However, existing work focuses on \doctable extraction over single flat tables~\cite{tang2024struc,DBLP:conf/emnlp/DengC00FZYS24,DBLP:conf/emnlp/Jiao0LZOJ023,liang2026long},
while the \docdb setting, which requires constructing multi-table relational databases with inter-table dependencies and integrity constraints from documents, remains largely unexplored.

\textbf{Existing Benchmarks.}
Prior benchmarks largely repurpose table-to-text datasets such as
Rotowire~\cite{wiseman2017challenges} and Wiki40B~\cite{guo2020wiki}, which
support only trivial extraction reducible to direct text replication.
LiveSum~\cite{DBLP:conf/emnlp/DengC00FZYS24} covers limited domains, while
synthetic benchmarks InstructIE~\cite{DBLP:conf/emnlp/Jiao0LZOJ023} and
StructText~\cite{DBLP:journals/corr/abs-2507-21340} suffer from simplified
schemata and trivial text--tuple alignments. Meanwhile, SemBench~\cite{lao2025sembench} and UDA-Bench~\cite{subramaniyaswamy2015unstructured}
target unstructured document analysis but lack fine-grained capability
evaluation.
More recent DTBench~\cite{dtbench2026} introduces capability-aware evaluation for document-to-table extraction, but remains limited to flat-table extraction without cross-table relational construction. SQUiD~\cite{DBLP:conf/emnlp/SadiaYXCC25} further explores text-to-relational database generation with multi-table schema synthesis, yet mainly evaluates logical relational view recovery from simple descriptions rather than realistic \docdb task requiring cross-table reasoning and database-level consistency.
These limitations motivate \ours, the first benchmark for systematic LLM
evaluation on document-to-database extraction.
\vspace{-0.5em}
\section{Conclusion}
\label{sec:conclusion}
\vspace{-.5em}


We introduce \ours, a novel benchmark advancing information extraction from isolated flat tables to relationally faithful document-to-database construction. To overcome manual annotation bottlenecks, we propose a controllable reverse-synthesis pipeline grounded in a comprehensive taxonomy of intra-table and inter-table capabilities. Extensive evaluations reveal that while state-of-the-art LLMs excel at localized entity extraction, they fundamentally struggle with global relational alignment, multi-hop structural reasoning, and adherence to database integrity constraints. \ours provides a rigorous and realistic testbed for advancing reliable, schema-compliant, and auditable LLM-based data systems, with further dataset expansion underway.


\section*{Limitations}

While \ours provides broad coverage across diverse schemas and domains, several aspects remain open for further exploration. The current DB2Doc synthesis pipeline relies on LLM-based generation, making human verification important for ensuring the quality and realism of synthesized documents. Beyond synthesis quality, we also plan to expand the benchmark by incorporating more domain-specific databases and broader real-world sources, strengthening its coverage and diversity.


\bibliographystyle{abbrv}
\bibliography{reference}


\appendix

\section{Dataset Collection and {Processing}}
\label{appendix:data_collection}
\vspace{-1em}

To establish a relationally faithful ground truth for \ours, we curated a collection of complex relational databases from two predominant text-to-SQL benchmarks: BIRD~\cite{li2023can} and Spider~\cite{yu2018spider}. The collection process followed a multi-stage pipeline designed to maximize domain diversity and structural complexity.

\subsection{Database Selection and Sourcing}
\vspace{-0.5em}
We first screen the database pools from BIRD and Spider, selecting databases that are suitable for document-to-database evaluation. 
Specifically, we prioritize databases with at least three interconnected tables, non-trivial schemas involving multiple entity types, relationship tables, and foreign-key dependencies, as well as diverse cross-table dependencies.
To ensure broad coverage, the final collection spans seven domains, including education, finance, healthcare, university, sports, transportation, and others.
Unlike conventional benchmarks that focus on flat-table extraction, our selection favors databases with rich entity-relationship (ER) structures and explicit relational dependencies, enabling the evaluation of cross-table reasoning and database construction capabilities.

\subsection{Data Cleansing and Schema Refinement}
\vspace{-0.5em}

\begin{figure*}[t!]
    \centering

    \includegraphics[width=1\textwidth]{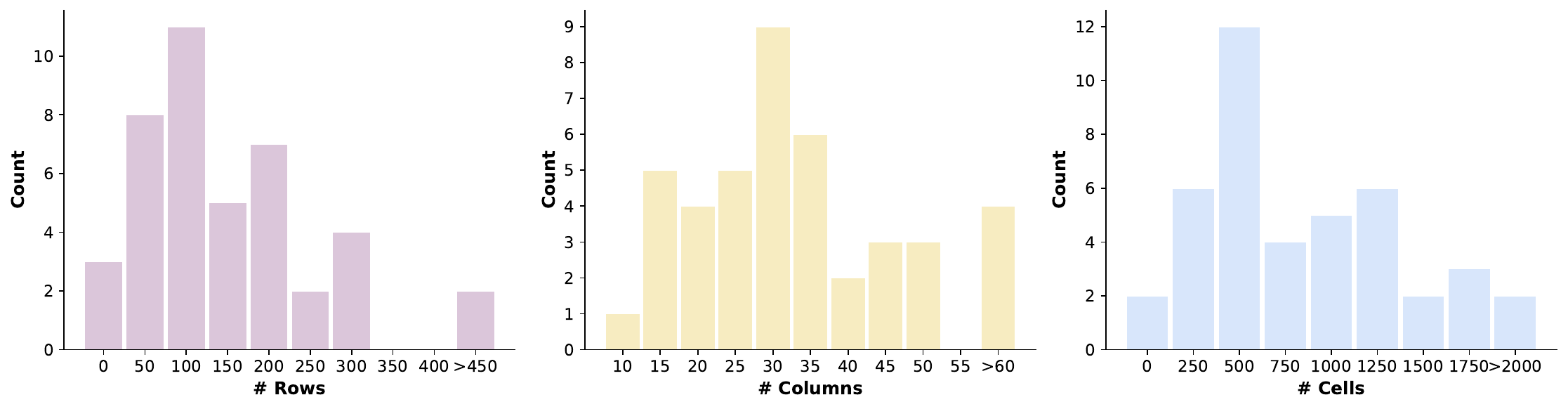}
    
    \caption{Distribution of database scale in \ours in terms of rows, columns, and cells.}
    \vspace{-1em}
    \label{fig:database_statistic}
\end{figure*}

The raw databases underwent several refinement steps to ensure they were suitable for document synthesis:
\begin{itemize}[leftmargin=*, itemsep=2pt]
\item \textbf{Schema Standardization:}
We normalize attribute names, unify schema representations, and ensure that primary and foreign keys are explicitly defined. This step reduces schema ambiguity and facilitates reliable DB-to-document synthesis.

\item \textbf{Tuple Sampling and Instance Construction:}
Since using all database tuples may produce excessively long documents and dilute evidence density, we sample representative tuples while preserving entity coverage, inter-table relationships, and referential integrity. 
The resulting instances remain complete relational databases rather than isolated table fragments, yielding 117 entity tables and 132 relationship tables with 7,341 rows and 41,935 cells. Fig.~\ref{fig:database_statistic} further illustrates the distribution of database scales in terms of rows, columns, and cells.

\item \textbf{Integrity Verification:}
We perform SQL-based sanity checks on the sampled databases to ensure that all instances satisfy relational constraints, including primary-key uniqueness, foreign-key referential integrity, and attribute-level validity. 
We further verify schema consistency, table connectivity, and tuple completeness to ensure that sampled instances preserve the original relational structures without introducing broken dependencies or isolated fragments.
Only verified database instances are retained for subsequent document synthesis.

\end{itemize}  

\vspace{-.5em}
\subsection{Domain-Specific Style Grounding}
Although databases from BIRD~\cite{li2023can} and Spider~\cite{yu2018spider} provide diverse domain schemas and structured facts, they are not designed to capture realistic document styles and writing conventions. 
To improve document realism, we augment the structured data with domain-specific narrative styles by collecting reference documents from LongDocURL~\cite{deng2025longdocurl}, DocBench~\cite{zou2025docbench}, MMLongBench~\cite{ma2024mmlongbench}, and MMDocRAG~\cite{dong2025benchmarking}. 
These references cover diverse document formats, ranging from academic papers and financial reports to posters and restaurant menus, serving as stylistic anchors for the synthesis pipeline. 
They enable the generation of authentic domain-specific documents while strictly preserving the factual constraints of the underlying databases.

\subsection{Verification Procedure.}
\label{ssec: appendix_verfication}
\vspace{-0.5em}

Although our dataset is generated through a multi-agent synthesis pipeline, we incorporate a rigorous verification procedure to ensure the quality and reliability of the synthesized documents.

First, our framework includes a dedicated \textbf{Validator Agent} that performs automatic verification from two perspectives. At the fine-grained level, it checks whether generated sentences and paragraphs faithfully cover the information contained in the source databases, preventing missing or incomplete evidence. At the database reconstruction level, it evaluates whether the complete database can be accurately recovered from the generated document. Any detected inconsistencies or missing information trigger an iterative repair process to refine the generated documents. Detailed verification and repair procedures are provided in Appendix~\ref{ssec:appendix_validor}.

Second, we conduct human verification on the generated documents. We recruit seven PhD candidates in computer science, with each reviewer responsible for one specific domain. Each reviewer evaluates the synthesized documents from two aspects: (1) whether the document style and content align with realistic domain-specific documents encountered in practice; and (2) whether the underlying database information can be reliably reconstructed from the generated document. 
Any cases that fail these criteria, such as unrealistic document patterns, missing database evidence, or inconsistent entity relationships, are filtered out to ensure the quality and reliability of the final dataset.

\subsection{Dataset Visualization.}
\vspace{-0.5em}

\begin{figure*}[t!]
    \centering

    \includegraphics[width=1\textwidth]{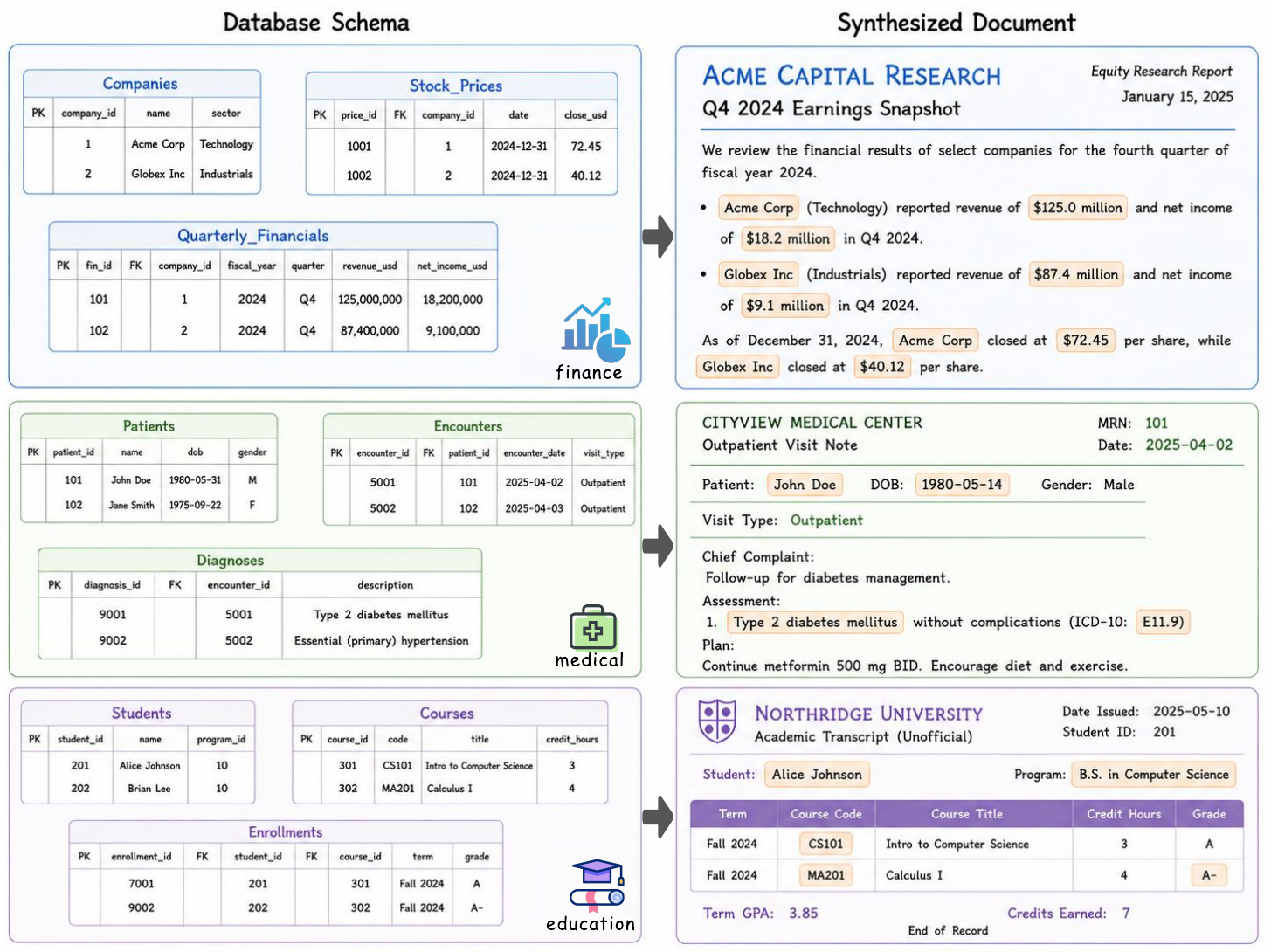}
    
    \caption{Representative examples from the finance, healthcare, and education domains. Each case pairs a multi-table relational schema with a domain-specific synthesized document, where colored links illustrate the grounding between database records and document evidence.}
    \vspace{-1em}
    \label{fig:dataset_visualization}
\end{figure*}

Figure~\ref{fig:dataset_visualization} presents representative examples from the finance, healthcare, and education domains, each pairing a multi-table relational schema with a realistic domain-specific document. For example, in the finance case, company information is stored in the \texttt{Companies} table, quarterly revenue and net income in \texttt{Quarterly\_Financials}, and closing prices in \texttt{Stock\_Prices}. The synthesized equity research report integrates these distributed records into statements such as ``Acme Corp reported revenue of \$125.0 million and net income of \$18.2 million,'' while linking them through the shared \texttt{company\_id}. Recovering the database therefore requires not only extracting individual values, but also resolving entity references, normalizing expressions such as \texttt{125,000,000} and ``\$125.0 million,'' and correctly assigning evidence to multiple related tables. Similar examples across the other domains illustrate the diversity of schema structures, document styles, and cross-table dependencies covered by \ours.

\section{Details of dual-level Taxonomy}
\label{sec:app_taxonomy}
\vspace{-.5em}

\begin{figure*}[t!]
    \centering

    \includegraphics[width=1\textwidth]{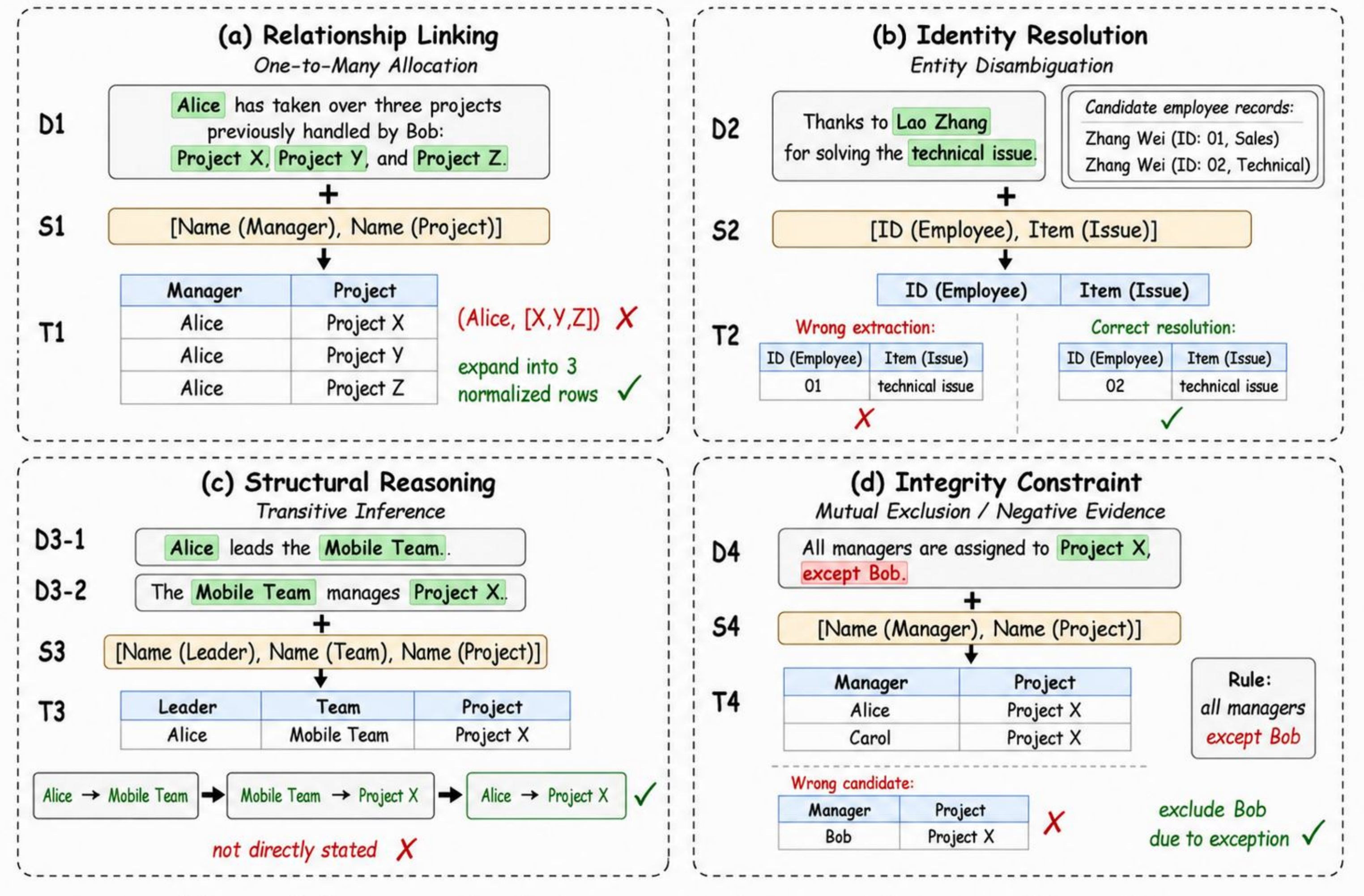}
    
    \caption{Challenging examples of relation extraction that require different inter-table capabilities.}
    \vspace{-1em}
    \label{fig:taxonomy_visualization}
\end{figure*}

Pillar~2 inter-table capabilities are organized into the following four types:

\stitle{Relationship Linking (RL).} 
This capability is required when natural language narratives must be parsed and mapped into structured relational tuples, accurately aligning multiple entities across tables based on cardinality and business rules. We categorize this linking into three subcategories.

\underline{\textit{One-to-Many Allocation.}} 
The document provides a compressed statement (\eg a manager taking over multiple projects), requiring the model to expand it into multiple normalized relationship rows that preserve the underlying 1:N or N:1 cardinality rather than collapsing them.
As illustrated in Fig.~\ref{fig:taxonomy_visualization} (a), a statement assigning Alice to Projects X, Y, and Z must be expanded into three separate relationship tuples.

\underline{\textit{Multi-Entity Binding.}} 
The document describes intertwined N-ary relations (\eg distinct suppliers delivering different parts to various projects) in a dense narrative, requiring exact triplet or multi-entity alignment without mismatching pairs or hallucinating spurious links based on text proximity.

\underline{\textit{Conditional Linkage.}} 
Requires evaluating existing table attributes against implicit business rules to automatically establish foreign key relations, even without explicit entity name mentions.

\stitle{Identity Resolution (IDR).} 
This capability is required when entity references within the document are ambiguous, pronominal, or use aliases, necessitating cross-referencing with table attributes or surrounding context to ground the mention to a unique, exact database record.

\underline{\textit{Entity Disambiguation.}} 
The document uses an ambiguous alias or name that matches multiple database records, requiring contextual clues and schema attributes to identify the correct entity.
As shown in Fig.~\ref{fig:taxonomy_visualization} (b), the alias ``Lao Zhang'' must be resolved to the employee in the technical department based on the surrounding ``technical issue'' context.

\underline{\textit{Coreference Resolution.}} 
The target entity is referred to using pronouns, role titles, or implicit aliases scattered across sentences, requiring cross-sentence contextual reasoning to correctly resolve these coreferences to the anchor entity and populate the corresponding relationship fields.

\stitle{Structural Reasoning (SR).} 
This capability is required when relational links cannot be directly extracted from a single text span but must be logically deduced by chaining multiple implicit relations, table structures, or temporal events.

\underline{\textit{Transitive Inference.}} 
The document and existing tables provide linked but disjointed relational pairs (\eg entity A is linked to B, and B is linked to C), requiring the deduction of implicit transitive relations to populate the target schema.
Fig.~\ref{fig:taxonomy_visualization} (c) illustrates this case: ``Alice leads the Mobile Team'' and ``the Mobile Team manages Project X'' must be composed to recover the implicit relation between Alice and Project X.

\underline{\textit{Global Aggregation.}} 
The attributes required to form a complete multi-dimensional relationship are scattered across disjointed text spans, requiring multi-hop reading comprehension to logically chain these spans and assemble all necessary foreign keys into a single valid record.

\underline{\textit{Dynamic Change.}} 
The document narrative involves cross-table event timelines, sequential actions, or entity state changes over time (e.g., completing a prerequisite course before a new enrollment), requiring reasoning over these temporal dynamics to determine the final, valid relationship status.

\stitle{Integrity Constraint (IC).} 
This capability is required when the extraction of relationships is governed by negative evidence, exceptions, or boundary rules, demanding that the model suppress naive co-occurrence extraction and strictly adhere to database integrity.

\underline{\textit{Null Relation Extraction.}} 
The document provides explicit negative evidence, such as cancelled events or unmet prerequisites, requiring the model to identify an empty relation state (\eg outputting \texttt{NULL}) and avoid hallucinating links from superficial keyword co-occurrence.

\underline{\textit{Mutual Exclusion.}} 
The document specifies broad relationships alongside explicit exceptions or negative constraints, requiring the model to infer relationship boundaries and exclude invalid relational tuples.
As illustrated in Fig.~\ref{fig:taxonomy_visualization} (d), although all managers are generally assigned to Project X, the explicit exception ``except Bob'' requires suppressing the tuple $(\text{Bob}, \text{Project X})$.

\section{Experiment Setup}
\vspace{-.5em}

\subsection{Metric Design Principles: Cell Alignment for Evaluation}
\label{ssec:app_metric}
\vspace{-0.5em}

Naive sequential matching aligns predicted rows to ground-truth rows by
position, which is fragile in two ways. Cross-Tuple Hijacking: a model extracts the correct value but assigns it to the wrong row, penalizing an otherwise valid prediction. Cascading Misalignment: a single missed or inserted row shifts all subsequent alignments, inflating both false-positive and false-negative counts. To address both failure modes, we adopt a \textbf{Global Maximum Weight Matching} strategy over a bipartite graph of ground-truth and predicted rows.

Notably, We do not use primary keys as alignment anchors because database-specific identifiers, particularly surrogate keys, are often absent from real-world documents and must be generated during database construction. Different models may therefore assign different identifiers to otherwise equivalent records. Primary-key-based matching would conflate arbitrary identifier generation with extraction correctness. We instead align rows using shared non-key attributes and evaluate key validity separately through database-integrity metrics.

Let $T^* = \{r^*_1, \ldots, r^*_m\}$ and $T = \{r_1, \ldots, r_n\}$ denote
the ground-truth and predicted row sets, respectively. Matching proceeds in
four stages.

\textbf{Stage 1: Global Scoring Matrix.}
Let $\mathcal{K}$ be the shared column names between $T^*$ and $T$. For each
pair $(r^*_i, r_j)$, we compute a composite similarity score:
\[
s(r^*_i,\, r_j) \;=\; \frac{1}{|\mathcal{K}|} \sum_{k \in \mathcal{K}}
\mathrm{sim}\!\left(r^*_i[k],\, r_j[k]\right),
\]
where $\mathrm{sim}(\cdot,\cdot)$ returns $1$ for exact numeric equality after normalization and uses normalized Levenshtein similarity for textual values:
\[
\mathrm{sim}_{\mathrm{text}}(a,b)
=
1-
\frac{
d_{\mathrm{Lev}}(\mathrm{norm}(a),\mathrm{norm}(b))
}{
\max\bigl(|\mathrm{norm}(a)|,|\mathrm{norm}(b)|\bigr)
},
\]
where $d_{\mathrm{Lev}}$ denotes the Levenshtein edit distance and
$\mathrm{norm}(\cdot)$ applies lowercasing, whitespace normalization, and
punctuation normalization. Missing or incompatible values receive a score of
$0$, yielding a global $m \times n$ scoring matrix.

\textbf{Stage 2: Global Sorting.}
All pairs with $s(r^*_i, r_j) \geq \tau$ ($\tau = 0.3$) are sorted globally
in descending order, ensuring high-confidence pairs are resolved first and
cannot be preempted by lower-scoring candidates.

\textbf{Stage 3: Double-Locking Assignment.}
Pairs are assigned greedily from the ranked list. Once $(r^*_i, r_j)$ is
matched, both indices are locked and excluded from further assignment,
enforcing strict one-to-one mapping and preventing score inflation via
duplicate row generation.

\textbf{Stage 4: Cell-Level Settlement.}
Within each matched pair, scoring proceeds cell-by-cell as defined in
Section~\ref{sec:experiments}. Unmatched ground-truth rows contribute all
non-empty cells to the missing count; unmatched predicted rows
contribute to the hallucination count, ensuring every cell is
accounted for exactly once.

\subsection{Prompts for LLM Score.}
\label{ssec: app_llmscore}
\vspace{-0.5em}

\begin{tcolorbox}[
    colback=gray!5,     
    colframe=gray!75,   
    title=\textbf{LLM-as-a-Judge: Database Extraction Evaluation Prompt},
    fonttitle=\small\bfseries,
    fontupper=\small,    
    arc=2pt,            
    outer arc=2pt,
    left=2mm, right=2mm, top=2mm, bottom=2mm, 
    enhanced,
    breakable           
]
You are a database evaluation expert. Please evaluate the quality of database tables extracted from documents.

\vspace{0.5em}
\textbf{Evaluation Dimensions:}
\begin{itemize}[leftmargin=1.5em, nosep]
    \item \textbf{Accuracy:} Semantic consistency, exact numeric matching (including decimal places), and string similarity $\ge$ 90\%.
    \item \textbf{Completeness:} Presence of all key information from the standard answer and structural integrity.
    \item \textbf{Standardization:} Compliance with Schema definitions and data type/format consistency.
\end{itemize}

\vspace{0.5em}
\textbf{Scoring Constraints:}
\begin{itemize}[leftmargin=1.5em, nosep]
    \item Numeric values must be \textit{exactly equal}; string similarity below 90\% is a mismatch.
    \item Type mismatches are considered serious errors.
    \item Total score ranges from 0 to 100; a perfect match equals 100 points.
\end{itemize}

\vspace{0.5em}
\textbf{Output Format:} 
Provide a comprehensive explanation ($\le$ 100 words), then output the score as: \\
\texttt{Evaluation evidence: [explanation] Rating: [[score]]}
\end{tcolorbox}

\section{More \ours Evaluation}
\label{sec:appendix_further_result}

\subsection{Database Integrity Analysis.}
\vspace{-0.5em}

\begin{figure*}[b!]
    \centering
    \includegraphics[width=\textwidth]{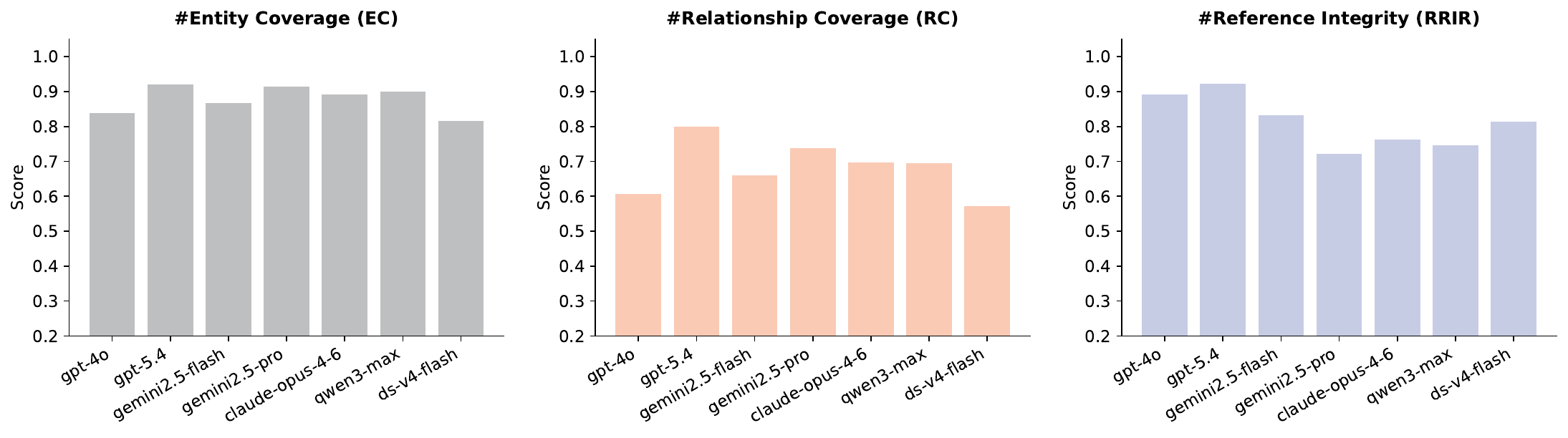}
    \caption{Database-level performance across Entity Coverage, Relation Coverage, and Reference Integrity Rate.}
    \vspace{-2em}
    \label{fig:db_integrity}
\end{figure*}

\textbf{Metric.}
Beyond attribute accuracy, we evaluate whether models can construct structurally complete and constraint-consistent databases.
We measure \emph{Entity Coverage} and \emph{Relation Coverage} to assess the recovery of entities and inter-entity relationships, respectively.
We further report \emph{Reference Integrity Rate}, which measures the proportion of foreign-key references that correctly point to existing entities, reflecting the validity and consistency of the constructed database.

\textbf{Result.}
Database-level results show that reconstruction requires both structural coverage and valid cross-table references. Entity Coverage exceeds Relation Coverage by 12.17--24.52 points across all models, highlighting the greater difficulty of recovering inter-table relationships. Coverage also does not guarantee validity: Gemini-2.5-Pro achieves strong coverage but the lowest RRIR, revealing frequent cross-table reference errors. GPT-5.4 leads all three metrics, demonstrating the strongest ability to recover relational structures while preserving database consistency. 

\vspace{-.5em} 
\begin{findingbox}[attach title to upper,after title={.\ }]{Finding 4} Database construction requires more than accurate extraction: maintaining relational consistency and valid entity references remains a critical challenge for LLMs. \end{findingbox} 
\vspace{-.5em}

\subsection{Full Results of Different Domains}
\vspace{-1em}

\begin{table*}[htbp]
\centering
\small
\caption{Results on \textsc{\ours} across various domains. \emph{Ver./Size} shows model version or size.}
\renewcommand\tabcolsep{3.1pt}
\renewcommand\arraystretch{1.45}

\label{tab:main_results}
\resizebox{\linewidth}{!}{%
\begin{tabular}{lc|ccccccc|c}
\toprule
\multirow{2}{*}{\textbf{Methods}} 
& \multirow{2}{*}{\makecell[c]{\textbf{Ver.}\\ \textbf{/Size}}} 
& \multicolumn{7}{c|}{\textbf{Domain}} 
& \multirow{2}{*}{\textbf{Overall $\mathrm{Acc.}$}} \\ 
\cmidrule(lr){3-9}
& & \textbf{Edu.} & \textbf{Fin.} & \textbf{Med.} & \textbf{Uni.} & \textbf{Sports} & \textbf{Trans.} & \textbf{Others} & \\
\midrule

\rowcolor{greyblue} \multicolumn{10}{c}{\emph{~~~~~~Open-source Models}}  \\ \\[-1.1em]
Qwen2.5 & \texttt{14B-Ins} & 13.25 & 34.21 & 37.35 & 36.44 & 18.16 & 47.31 & 34.82 & 32.97 \\ 
LLaMA-3.1 & \texttt{70B-Ins} & 10.94 & 13.18 & 16.29 & 21.54 & 27.86 & 7.30 & 8.16 & 14.29 \\ 
Qwen2.5 & \texttt{72B-Ins} & 30.41 & 34.79 & 40.98 & 47.73 & 31.87 & 38.80 & 29.98 & 36.07 \\ 

\midrule
\rowcolor{greyblue} \multicolumn{10}{c}{\emph{~~~~~~Proprietary Models}}  \\ \\[-1.1em]
GPT-4o & \texttt{-} & 47.89 & 57.49 & 54.20 & 72.32 & 62.75 & 68.02 & 51.11 & 59.05 \\ 
GPT-5.4 & \texttt{-} & 67.67 & \textbf{68.65} & 66.68 & 74.78 & \textbf{82.24} & 85.22 & \textbf{78.49} & \textbf{75.25} \\ 
Gemini 2.5 & \texttt{Flash} & 57.63 & 64.04 & 41.00 & 73.13 & 71.22 & 78.95 & 64.57 & 65.99 \\ 
Gemini 2.5 & \texttt{Pro} & \textbf{69.81} & 65.46 & \textbf{76.53} & 79.75 & 69.67 & 74.78 & 67.94 & 70.81 \\ 
Claude Opus & \texttt{4-6} & 67.01 & 65.38 & 75.51 & \textbf{80.01} & 78.36 & \textbf{87.83} & 69.89 & 73.60 \\ 
Qwen3 & \texttt{Max} & 52.06 & 62.55 & 56.37 & 72.15 & 57.24 & 70.73 & 64.10 & 52.80 \\ 
DeepSeek-V4 & \texttt{Flash} & 53.66 & 59.39 & 48.23 & 55.60 & 57.79 & 73.56 & 51.80 & 57.06 \\ 
\bottomrule
\end{tabular}
}
\end{table*}

The results reveal a substantial performance gap between open-source and proprietary models. Among open-source models, Qwen2.5-72B achieves the highest overall accuracy of 36.07, but still trails GPT-5.4 by 39.18 points. GPT-5.4 obtains the best overall accuracy of 75.25 and leads in finance, sports, and other domains. Nevertheless, no single model dominates across all domains: Gemini 2.5 Pro performs best in education and medicine, while Claude Opus 4-6 leads in university and transportation, suggesting that different domains place distinct demands on schema interpretation, cross-table reasoning, and domain-specific information normalization.

Domain-level performance can be further understood from the normalized dataset statistics. Averaged across all evaluated models, transportation achieves the highest accuracy (63.25), followed by university (61.35), whereas education obtains the lowest accuracy (47.03). Education contains the largest number of rows per database (276.7), together with 8.0 tables and 4.0 relation tables on average, requiring models to recover a relatively large number of tuples while preserving cross-table associations. Finance is also challenging, but primarily because of its relational density rather than its absolute size: each database contains 6.2 tables and 3.9 relation tables on average, with the highest relation-to-entity ratio of 1.70. Medical databases present a different difficulty, having the largest average numbers of columns (66.3) and cells (1,174.0) per database, which increases the burden of broad attribute coverage. In contrast, although transportation documents are the longest on average (53,091 tokens), their databases contain only 3.6 tables and 1.2 relation tables per database, suggesting that document length alone is not the dominant source of difficulty. Overall, the results indicate that performance is more strongly influenced by the interaction of tuple volume, schema width, relation density, and implicit cross-table dependencies than by document or database scale in isolation.

\section{Implementation of \docdb}
\label{sec:appendix_implementation}
\vspace{-1em}

During the entire \textit{Controllable DB2Doc Reverse-Synthesis} process, we design specific prompts and rigorous verification protocols across four key agentic modules to ensure generation fidelity and relational complexity. 

Firstly, in the capability assignment stage, the \textbf{Labeling Agent} acts as a semantic router. We construct prompts to strategically assign fine-grained intra-table (Pillar 1) and inter-table (Pillar 2) capability labels, which directly control the downstream extraction difficulty. 
Secondly, during the evidence decomposition stage, the \textbf{Refiner Agent} processes the annotated tables into atomic evidence. To guarantee the mathematical and semantic rigor of this step, we implement a strict \textit{checklist-based verification} protocol. We design prompts that explicitly instruct the model to verify four critical dimensions: Value Correctness, Label Faithfulness, Schema Linkage, and Null Faithfulness.
Thirdly, to bridge the gap between discrete evidence and cohesive natural language, the \textbf{Writer Agent} executes the document generation. We adopt a context-aware generation paradigm where the prompt incorporates a sliding window of historical text and configurable stylistic profiles to emulate real-world document morphological complexity
Finally, in order to obtain a high-quality, hallucination-free corpus, the \textbf{Validator Agent} conducts a rigorous dual-validation pass. We construct prompts to evaluate Coverage Verification and Extraction Consistency. Documents that fail these checks trigger an iterative reject-and-rewrite mechanism.

\subsection{The prompt for Labeling Agent}
\vspace{-0.5em}

\definecolor{cBlue_6}{HTML}{2B4B80} 
\definecolor{cBlue_1}{HTML}{007AFF} 

\definecolor{titleBlue1}{HTML}{3D7197}
\definecolor{contentBlue1}{HTML}{2A5574}

\definecolor{titleBlue2}{HTML}{4F6E94}
\definecolor{contentBlue2}{HTML}{2A5574}

\noindent 
\begin{tcolorbox}[
    equal height group=pillarboxes,
    width=0.49\linewidth,
    nobeforeafter,
    colback=cBlue_1!2,
    colframe=titleBlue1,
    colbacktitle=titleBlue1,
    coltitle=white,
    title=\textbf{Pillar 1 Assignment Prompt},
    fonttitle=\bfseries\small,
    fontupper=\footnotesize,
    boxrule=0.4mm,
    arc=1.5mm,
    left=4pt, right=4pt, top=4pt, bottom=4pt,
    enhanced
]

\textbf{Goal:} Assign intra-table (Pillar 1) capability labels to target cells in the entity table.

\vspace{0.3em}
\textbf{Inputs:}
\begin{itemize}[leftmargin=*, nosep]
    \item \textbf{Input Table:} \{markdown\_table\}
    \item \textbf{Non-target Key Cols:} \{non\_target\_key\_columns\}
    \item \textbf{Attribute Schema:} \{attribute\_descriptions\}
    \item \textbf{Capability Definitions:} \{capability\_definitions\}
    \item \textbf{Target Label Ratio:} \{label\_ratio\}
\end{itemize}

\vspace{0.3em}
\textbf{Instructions:}
\begin{enumerate}[leftmargin=*, nosep]
    \item Assign labels that increase single-table extraction difficulty while preserving correctness.
    \item \textbf{Type-Strategy Compatibility (CRITICAL):} Independently analyze the semantic nature and data type of each attribute. Only assign strategies logically/mathematically suitable for the cell's value.
    \item A cell may have 0, 1, or multiple strategies (default is none).
    \item Assign creatively but ensure text remains fluent. Avoid over-complicating simple facts.
    \item DO NOT assign labels to Non-target Key Columns.
    \item For composite primary keys, use \texttt{", "} as separator.
    \item Label coverage: at least the target ratio of assignable cells must have non-empty labels.
    \item Return JSON only.
\end{enumerate}

\vspace{0.3em}
\textbf{Output JSON Schema:}\\
{\ttfamily
\{"assignments": \{"$<$pk\_1$>$": \{"$<$attr\_1$>$": []\} \} \}
}

\end{tcolorbox}
\hfill
\begin{tcolorbox}[
    equal height group=pillarboxes,
    width=0.49\linewidth,
    nobeforeafter,
    colback=cBlue_1!2,
    colframe=titleBlue2,
    colbacktitle=titleBlue2,
    coltitle=white,
    title=\textbf{Pillar 2 Assignment Prompt},
    fonttitle=\bfseries\small,
    fontupper=\footnotesize,
    boxrule=0.4mm,
    arc=1.5mm,
    left=4pt, right=4pt, top=4pt, bottom=4pt,
    enhanced
]

\textbf{Goal:} Assign inter-table (Pillar 2) capability labels to target cells in the relation table.

\vspace{0.3em}
\textbf{Inputs:}
\begin{itemize}[leftmargin=*, nosep]
    \item \textbf{Input Table:} \{markdown\_table\}
    \item \textbf{Table Schema Context:} \{relation\_context\}
    \item \textbf{Attribute Schema:} \{attribute\_descriptions\}
    \item \textbf{Capability Definitions:} \{capability\_definitions\}
    \item \textbf{Target Label Ratio:} \{label\_ratio\}
\end{itemize}

\vspace{0.3em}
\textbf{Label Assignment Constraints:}
\begin{itemize}[leftmargin=*, nosep]
    \item \texttt{IDR\_ED}, \texttt{RL\_O2M}, \texttt{RL\_MB}, and \texttt{RL\_MI} must each appear alone. Do NOT combine them.
    \item \texttt{IDR\_ED}, \texttt{RL\_O2M} requires that the same foreign key entity appears in $\ge$ 2 rows. Do NOT assign if unique.
\end{itemize}

\vspace{0.3em}
\textbf{Instructions:}
\begin{enumerate}[leftmargin=*, nosep]
    \item Assign labels that increase inter-table linking, topology, and structural extraction difficulty.
    \item \textbf{Topology-Strategy Compatibility (CRITICAL):} Analyze structural distribution. Only assign relational strategies logically suitable for the topology.
    \item Output ONLY row-level labels per relation row\_key. Do NOT output per-attribute labels.
    \item A row may have 0, 1, or multiple strategies (default is none).
    \item Top-level key MUST use the relation table PK string.
    \item Composite keys format: \texttt{$<$pk1$>$, $<$pk2$>$} (e.g., \texttt{12, 10}).
    \item Label coverage: at least the target ratio of rows must contain non-empty labels.
    \item Return JSON only.
\end{enumerate}

\vspace{0.3em}
\textbf{Output JSON Schema:}\\
{\ttfamily
\{"assignments": \{"$<$pk\_1$>$": [], "$<$pk\_2$>$": []\} \}
}

\end{tcolorbox}

The prompt above illustrates how the Labeling Agent acts as a semantic router to assign fine-grained intra-table (Pillar 1) and inter-table (Pillar 2) capability labels. It explicitly instructs the model to analyze data types and topological contexts, ensuring that specific reasoning challenges are injected logically to control downstream extraction difficulty while adhering to target label ratios.

\subsection{Refiner: Principle of Checklist}
~\label{ssec:app_checklist}
\vspace{-1em}

Given the annotated targets, the Refiner decomposes each target into atomic evidence that realizes its assigned capability labels, forming a global evidence pool for subsequent document generation.
When multiple labels are assigned, the evidence must jointly satisfy the corresponding intra-table and inter-table reasoning requirements.
To ensure the rigor of the synthesized evidence, we deploy an LLM-as-a-verifier~\cite{zheng2023judging} using a strict checklist-based protocol.
The verification assesses the generated evidence guidance across four critical dimensions:

\begin{itemize}[leftmargin=*, nosep]
\item \textbf{Value and Relation Correctness:}
Verifies that the guidance accurately reflects target cell values and inter-record relations, ensuring numerical correctness for calculated fields and allowing complete relational tuples to be unambiguously recovered.

\item \textbf{Label Faithfulness:}
Verifies that the guidance faithfully instantiates the assigned capability labels and rejects cases that omit or misrepresent the intended reasoning requirements.

\item \textbf{Schema Linkage:}
Ensures that the guidance is grounded in valid schema elements and preserves the intended structural dependencies among tables.
For relational tables, primary- and foreign-key references must be correctly identified and linked, so that participating entities can be unambiguously connected and the resulting evidence remains consistent with the target database schema.

\item \textbf{Data Integrity \& Null Faithfulness:}
Ensures that the guidance contains only information supported by the target database and does not introduce values from unrelated cells or records.
It also verifies that missing, unavailable, or non-applicable attributes remain uninstantiated, preventing the generation of unsupported values or spurious relational links.
\end{itemize}

On failure, the verifier acts as a critic, returning a specific rationale and actionable revision suggestions (e.g., ``Instruct the writer to explicitly mention both the supplier and project names''). This process iteratively refines the text until it passes all checks or reaches a retry limit.

\subsection{The prompt for Writer Agent}
\vspace{-0.5em}

\begin{tcolorbox}[
    colback=gray!5,     
    colframe=gray!75,   
    title=\textbf{Writer Agent Prompt}, 
    fonttitle=\small\bfseries,
    fontupper=\small,    
    arc=2pt,             
    outer arc=2pt,
    left=2mm, right=2mm, top=2mm, bottom=2mm, 
    enhanced,
    breakable            
]

\textbf{[System Role \& Objective]} \\
You are an expert Writer Agent operating within a parameterized document synthesis pipeline. \\
Your goal is to write ONE coherent, polished, and natural-sounding section of a larger document while perfectly preserving structured ground-truth facts.

\vspace{0.5em}
\textbf{[Generation Context]}
\begin{itemize}
    \setlength{\itemsep}{0pt}
    \item Previous Block Context (Tail Anchor): \{previous\_context\}
    \item Current Block ID: \{block\_id\} (Index: \{block\_index\} of \{total\_blocks\})
\end{itemize}

\vspace{0.5em}
\textbf{[Base Parameters]}
\begin{itemize}
    \setlength{\itemsep}{0pt}
    \item Global Target Document Length: \{document\_length\_tokens\} tokens
    \item Suggested Token Budget for THIS Block: \{target\_block\_tokens\} tokens
    \item Document Style \& Tone: \{document\_style\}
    \item Section Template Hint: \{section\_template\}
\end{itemize}

\vspace{0.5em}
\textbf{[Injection Protocols]}
\begin{itemize}
    \setlength{\itemsep}{0pt}
    \item Hard Cases Protocol:  \{hard\_cases\_protocol\}
    \item Controlled Noise Protocol:  \{noise\_protocol\}
    \item Representation Complexity Protocol:  \{complexity\_protocol\}
\end{itemize}

\vspace{0.5em}
\textbf{[Mandatory Ground Truth (Evidence)]} \\
You MUST explicitly embed the following structured facts. Do NOT drop any fact: 
\{facts\_with\_tags\}

\vspace{0.5em}
\textbf{[Execution Instructions]}
\begin{enumerate}
    \setlength{\itemsep}{0pt}
    \item FACTUAL FIDELITY: Embed all mandatory facts without altering their meaning.
    \item TAG PRESERVATION: Keep facts wrapped in their original tags (\eg \texttt{<rX\_cY>}).
    \item STYLISTIC ADAPTATION: Match the requested style, tone, and template.
    \item PROTOCOL COMPLIANCE: Strictly follow any active injection protocols; ignore empty ones.
    \item LENGTH \& PACING: Stay close to the \texttt{\{target\_block\_tokens\}} budget.
    \item SEAMLESS BRIDGING: Continue logically and smoothly from the previous context.
    \item FORMATTING: Use natural prose paragraphs. No bullets, markdown, or JSON.
    \item OUTPUT STRUCTURE: Start exactly with the Section Title, followed by the body.
    \item FINAL OUTPUT: Return ONLY the synthesized text, without conversational filler.
\end{enumerate}

\end{tcolorbox}

The prompt above demonstrates how the Writer Agent transforms serialized atomic evidence into cohesive, long-form natural language documents. It guides the model to employ a context-aware generation paradigm, seamlessly weaving discrete facts into fluent narratives while conforming to designated stylistic profiles and mimicking the morphological complexity of real-world corpora.

\subsection{The prompt for Validator Agent}
\label{ssec:appendix_validor}
\vspace{-0.5em}

\definecolor{cBlue_6}{HTML}{2b4b80} 
\definecolor{cBlue_1}{HTML}{007AFF} 

\definecolor{titleBlue1}{HTML}{3D7197}
\definecolor{contentBlue1}{HTML}{2A5574}

\definecolor{titleBlue2}{HTML}{4F6E94}
\definecolor{contentBlue2}{HTML}{2A5574}

\noindent
\begin{tcolorbox}[
    equal height group=promptboxes,
    width=0.49\linewidth,
    nobeforeafter,
    colback=gray!5,
    colframe=gray!75,
    colbacktitle=gray!75,
    coltitle=white,
    title=\textbf{Verification Agent Prompt},
    fonttitle=\small\bfseries,
    fontupper=\footnotesize,
    boxrule=0.4mm,
    arc=2pt,
    outer arc=2pt,
    left=2mm,
    right=2mm,
    top=2mm,
    bottom=2mm,
    enhanced
]

\textbf{Your Goal:} \\
Verify whether the generated text is faithful to the table content without fabrication.

\vspace{0.5em}
\textbf{Table Content:} \{markdown\_table\}
\vspace{0.3em}

\textbf{Generated Text:} \{generated\_text\}

\vspace{0.5em}
\textbf{Verification Checks:} \\
To ensure \textbf{Cell Extraction Consistency}, Key values stated in \texttt{generated\_text} must be extractable or inferrable from \texttt{Table Content}.

\vspace{0.5em}
\textbf{Output Format:} \\
Respond with ONLY a single, valid JSON object: \\
\vspace{0.3em}
{\ttfamily
\{ \\
\hspace*{1em}"ok": true/false, \\
\hspace*{1em}"errors": [ \\
\hspace*{2em}\{ \\
\hspace*{3em}"description": "Error description", \\
\hspace*{3em}"suggestion": "How to fix" \\
\hspace*{2em}\} \\
\hspace*{1em}] \\
\}
}
\vspace{0.3em}

\vspace{0.5em}
\textbf{FINAL INSTRUCTION:} Output ONLY the valid JSON object. If any check fails, set \texttt{ok} to false and include all issues in \texttt{errors}.

\end{tcolorbox}
\hfill
\begin{tcolorbox}[
    equal height group=promptboxes,
    width=0.49\linewidth,
    nobeforeafter,
    colback=gray!5,
    colframe=gray!75,
    colbacktitle=gray!75,
    coltitle=white,
    title=\textbf{Repair Agent Prompt},
    fonttitle=\small\bfseries,
    fontupper=\footnotesize,
    boxrule=0.4mm,
    arc=2pt,
    outer arc=2pt,
    left=2mm,
    right=2mm,
    top=2mm,
    bottom=2mm,
    enhanced
]
\textbf{Goal:} Repair a section of text that failed verification.

\vspace{0.5em}
\textbf{Input:}
\begin{itemize}
    \setlength{\itemsep}{0pt}
    \item \textbf{Original Text:}  \{content\}
    \item \textbf{Errors:}  \{errors\}
    \item \textbf{Required Facts:}  \{facts\}
\end{itemize}

\vspace{0.5em}
\textbf{Instructions:}
\begin{enumerate}
    \setlength{\itemsep}{0pt}
    \item Repair the section to fix the reported errors.
    \item Ensure all required facts remain correct.
    \item Keep the narrative flow natural.
    \item Preserve all provenance-style tags already present in the original text.
\end{enumerate}

\vspace{0.5em}
\textbf{Output Format:} \\
Return ONLY the repaired text content. No titles, no metadata, no markdown code blocks.

\end{tcolorbox}

The prompt above details the rigorous dual-validation protocol executed by the Validator Agent to guarantee the fidelity and relational consistency of the synthesized documents. It tasks the model with strictly evaluating coverage verification and extraction consistency, ensuring no evidence is omitted or hallucinated, which inherently serves as the trigger for our iterative reject-and-rewrite mechanism.

\section{Authenticity Verification}
\label{sec: appendix_authenticity}

\subsection{LLM \& Human Score}
\vspace{-0.5em}

\begin{table*}[t]
\centering
\caption{Evaluation rubrics for generated documents (1=worst, 5=best).}
\label{tab:document_evaluation_rubrics}
\small
\setlength{\tabcolsep}{6pt}
\renewcommand{\arraystretch}{1.45}

\begin{tabularx}{\textwidth}{
    >{\centering\arraybackslash}p{0.045\textwidth}
    >{\raggedright\arraybackslash}X
    >{\raggedright\arraybackslash}X
    >{\raggedright\arraybackslash}X
    >{\raggedright\arraybackslash}X
}
\toprule
\textbf{Score}
& \textbf{Lexical Richness}
& \textbf{Logical Consistency}
& \textbf{Textual Coherence}
& \textbf{Genre Fidelity} \\
\midrule

1
& \textbf{Repetitive:}\newline
  Minimal variety, robotic repetition
& \textbf{Fragmented:}\newline
  No connections, random claims
& \textbf{Incoherent:}\newline
  Difficult to follow, random jumps
& \textbf{Mismatch:}\newline
  Incorrect genre conventions 
\\

2
& \textbf{Limited:}\newline
  Simple vocabulary, narrow range
& \textbf{Weak:}\newline
  Loose transitions, vague reasoning
& \textbf{Poor flow:}\newline
  Jarring transitions, disconnected
& \textbf{Weak:}\newline
  Limited and inconsistent cues
\\

3
& \textbf{Acceptable:}\newline
  Adequate variety, standard usage
& \textbf{Acceptable:}\newline
  Clear order, basic signaling
& \textbf{Acceptable:}\newline
  Some awkward transitions
& \textbf{Plausible:}\newline
  Recognizable but generic style
\\

4
& \textbf{Versatile:}\newline
  Natural synonyms, precise terminology
& \textbf{Compelling:}\newline
  Strong arguments, coherent progression
& \textbf{Smooth:}\newline
  Clear progression, minor issues
& \textbf{Authentic:}\newline
  Consistent genre-specific conventions
\\

5
& \textbf{Sophisticated:}\newline
  Rich nuances, professional mastery
& \textbf{Rigorous:}\newline
  Flawless chain, seamless attribution
& \textbf{Seamless:}\newline
  Natural flow, effortless transitions
& \textbf{Professional:}\newline
  Realistic and polished genre presentation
\\

\bottomrule
\vspace{1em}
\end{tabularx}
\end{table*}

The synthesized documents in \textsc{\ours} should be fluent, logically coherent, and faithful to the conventions of real-world domain documents rather than appearing as artificial or template-based constructions. To evaluate document quality and authenticity, we adopt a four-dimensional framework combining LLM-as-a-judge with human evaluation. As shown in Table~6, the framework assesses \textit{lexical richness}, \textit{logical consistency}, \textit{textual coherence}, and \textit{genre fidelity}, each assessed using a five-point rubric, following a similar strategy of previous work (\eg LLMs4Synthesis~\cite{babaei2024llms4synthesis} and DTBench~\cite{dtbench2026}). We use DeepSeek-V3.2 as the automatic evaluator. In addition, five graduate students with research experience in document intelligence independently assess 50 randomly sampled documents using the same rubrics. The LLM and human evaluation results in Table~\ref{tab:authenticity} consistently demonstrate the high linguistic quality and domain authenticity of the synthesized documents.

\subsection{Real Case Comparison}
\vspace{-1em}

\begin{figure*}[t!]
    \centering

    \includegraphics[width=1\textwidth]{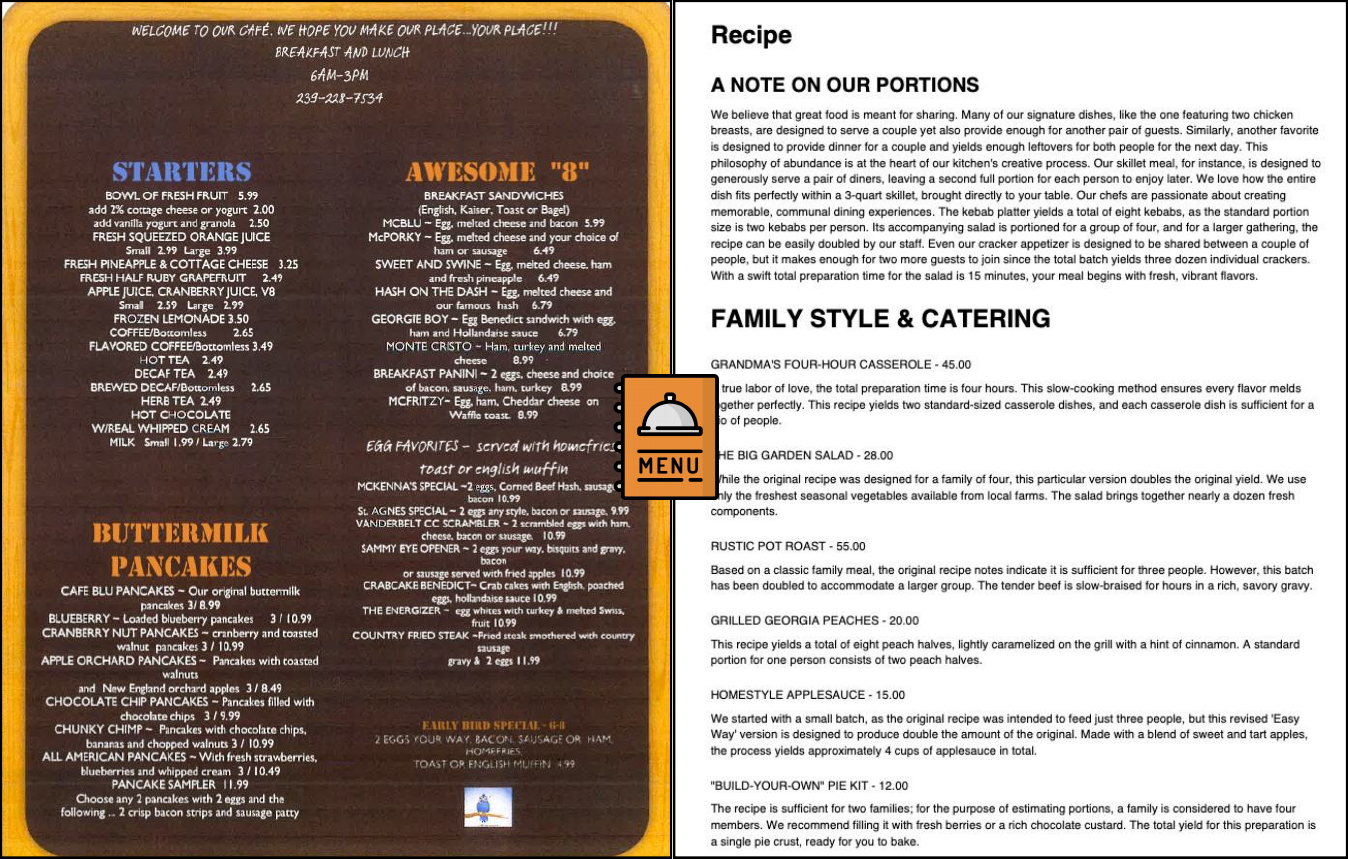}
    
    \caption{menu case}
    \vspace{-1em}
    \label{fig:menu_case}
\end{figure*}

\begin{figure*}[t!]
    \centering

    \includegraphics[width=1\textwidth]{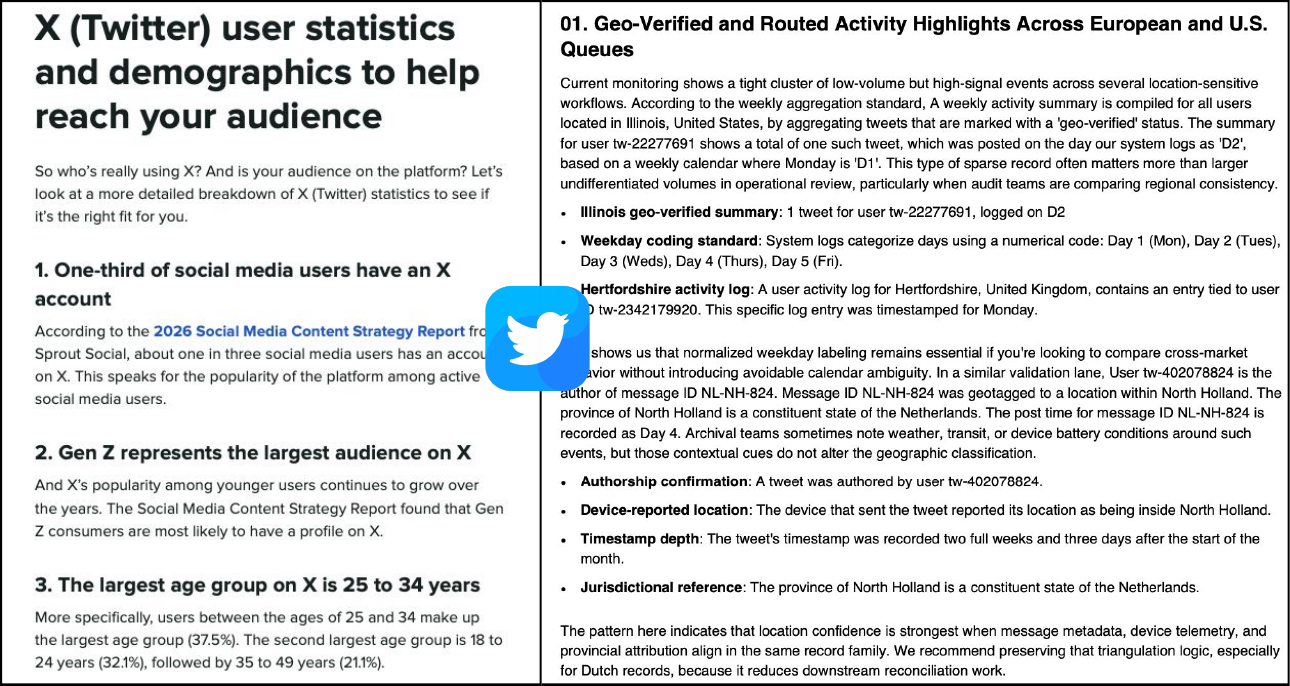}
    
    \caption{twitter case}
    \vspace{-1em}
    \label{fig:twitter_case}
\end{figure*}

Fig.~\ref{fig:menu_case}~--~\ref{fig:twitter_case} provide qualitative comparisons between synthesized and real-world documents across diverse domains, including finance, aviation, media, and restaurant menus, further demonstrating the quality and realism of our synthesis pipeline.

\begin{figure*}[t!]
    \centering

    \includegraphics[width=1\textwidth]{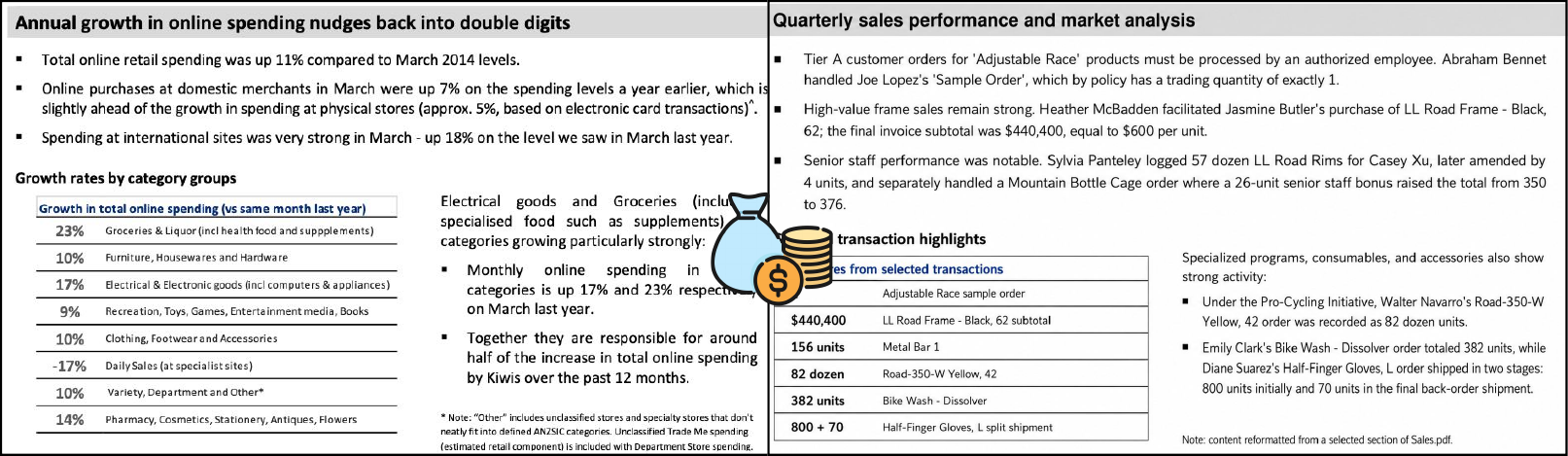}
    
    \caption{finance case}
    \vspace{-1em}
    \label{fig:finance_case}
\end{figure*}

\begin{figure*}[t!]
    \centering

    \includegraphics[width=1\textwidth]{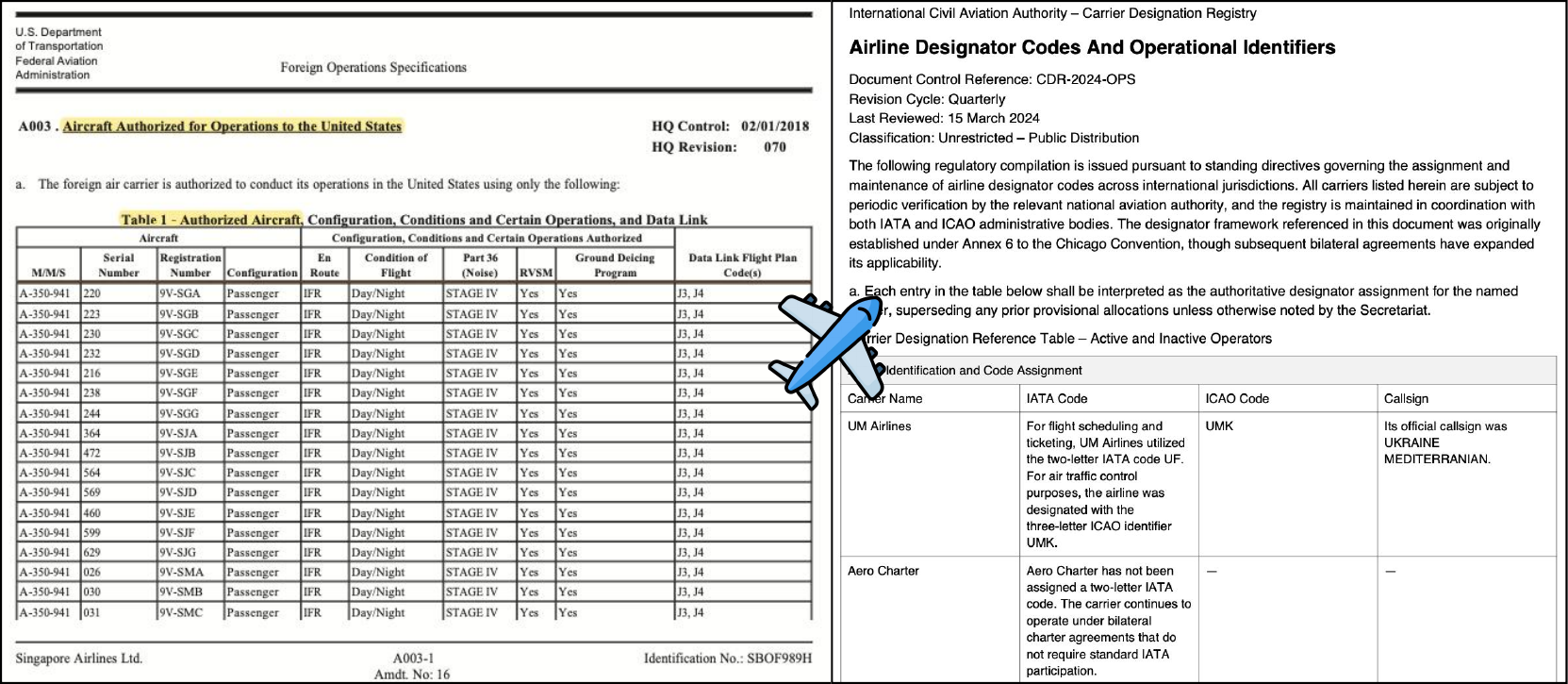}
    
    \caption{airline case}
    \vspace{-1em}
    \label{fig:finance_case}
\end{figure*}

\end{document}